\PassOptionsToPackage{table}{xcolor}
\documentclass[a4paper,fleqn]{cas-sc}

\usepackage[numbers]{natbib}

\usepackage[utf8]{inputenc} 
\usepackage{url}            
\usepackage{nicefrac}       
\usepackage{microtype}      
\usepackage{siunitx}
\usepackage{float}
\usepackage{dsfont}
\usepackage{subcaption}
\usepackage{pifont}
\usepackage{threeparttable}

\ExplSyntaxOn
\cs_set:Npn \__make_tbl_caption:nn #1#2
{
  \l_tbl_align_tl
  \skip_vertical:N \l_tbl_abovecap_skip
  {\parbox{\linewidth}
    {\rightskip=0pt\sffamily\small\textbf{\color{scolor}#1}\par#2\par\vskip4pt }}
  \skip_vertical:N \l_tbl_belowcap_skip
}
\ExplSyntaxOff

\newcommand{\minisection}[1]{\vspace{0.0in} \noindent {\bf #1}}

\newcommand{\eg}{\textit{eg.}}

\newcommand{\ie}{\textit{ie.}}
\definecolor{best}{RGB}{0,120,0}

\newcommand{\method}{R2RI\xspace}

\newcommand{\mylist}[1]{%
  \begingroup
    \spaceskip=\fontdimen2\font plus \fontdimen3\font minus \fontdimen4\font
    \xspaceskip=\fontdimen7\font\relax
    \ttfamily
    #1%
  \endgroup
}

\definecolor{nored}{RGB}{190, 0, 0}    
\definecolor{yesgreen}{RGB}{0, 150, 0} 

\begin{document}
\let\WriteBookmarks\relax
\def\floatpagepagefraction{1}
\def\textpagefraction{.001}

\shorttitle{R2RI: A Multi-View Event and RGB Dataset for Robot-to-Robot Interaction}

\shortauthors{G. Magrini et al.}

\title [mode = title]{R2RI: A Multi-View Event and RGB Dataset for Robot-to-Robot Interaction}


\author[unifi]{Gabriele Magrini}
\cormark[1]
\fnmark[1]
\ead{gabriele.magrini@unifi.it}
\credit{Conceptualization, Methodology, Software, Data curation, Investigation, Writing -- original draft}

\author[unimore]{Riccardo Catalini}
\fnmark[1]
\credit{Conceptualization, Methodology, Software, Data curation, Investigation, Writing -- original draft}

\author[unisi]{Federico Becattini}
\credit{Conceptualization, Funding acquisition, Supervision, Validation, Writing -- original draft, Writing -- review and editing}

\author[unimore]{Guido Borghi}
\credit{Conceptualization, Supervision, Writing -- original draft, Writing -- review and editing}

\author[unifi]{Pietro Pala}
\credit{Resources, Funding acquisition, Supervision, Writing -- review and editing}

\author[unimore]{Roberto Vezzani}
\credit{Resources, Funding acquisition, Supervision, Writing -- review and editing}

\author[unifi]{Lorenzo Seidenari}
\credit{Conceptualization, Methodology, Supervision, Writing -- review and editing}

\affiliation[unifi]{organization={Department of Information Engineering, University of Florence},
            city={Florence},
            country={Italy}}

\affiliation[unimore]{organization={Department of Engineering ``Enzo Ferrari'', University of Modena and Reggio Emilia},
            city={Modena},
            country={Italy}}

\affiliation[unisi]{organization={Department of Information Engineering and Mathematics (DIISM), University of Siena},
            city={Siena},
            country={Italy}}

\cortext[1]{Corresponding author}

\fntext[1]{These authors contributed equally to this work.}

\begin{abstract}
Understanding and modeling interactions between autonomous agents is a fundamental challenge in robotics, with broad implications for collaborative systems, social robotics, and human-robot coexistence. Although the study of robot interactions has emerged as a compelling research direction, progress has been severely hampered by the absence of large-scale benchmarks. In this paper, we introduce Robot-to-Robot Interaction (R2RI), the first dataset specifically designed to address the Robot-Robot Interaction (RRI) task. R2RI consists of different humanoid robots and realistic interactions modeled on real human social behaviors. Complementary viewpoints are available, \textit{i.e.}, an egocentric perspective from each robot's onboard sensors, and an exocentric perspective from external fixed cameras, thus enabling rich spatial and contextual understanding of the interaction dynamics. 
The dataset comprises more than $6.5$M frames and $\approx5000$ videos at $120$ fps, including Event and RGB domains. We investigate pros and cons of each domain, comparing state-of-the-art approaches for a number of key sensing and interaction based tasks.
We publicly release the dataset and its annotations for all tasks and modalities at \url{https://github.com/MagriniGabriele/R2RI}.
\end{abstract}



\maketitle

\section{Introduction}\label{sec1}
As autonomous robotic systems are increasingly deployed in shared environments - factories, hospitals, public spaces, and homes - endowing them with analogous interaction-aware capabilities becomes not merely desirable, but essential~\cite{peshkin2001cobot,bicchi2008safety,weiss2011exploring}. However, while substantial research has been devoted to Human-Robot Interaction (HRI)~\cite{kolbeinsson2019foundation} and, more broadly, to the perception of human actions and intentions, the specific problem of Robot-Robot Interaction (RRI) has received comparatively little systematic attention. Yet this is a setting with profound practical relevance: multi-robot systems must coordinate, negotiate space, communicate intent, and avoid conflict in real time, often without explicit communication channels and under strict latency constraints~\cite{simoni20253d,dautenhahn2011new}.
\begin{figure}
    \centering
    \includegraphics[width=.9\linewidth]{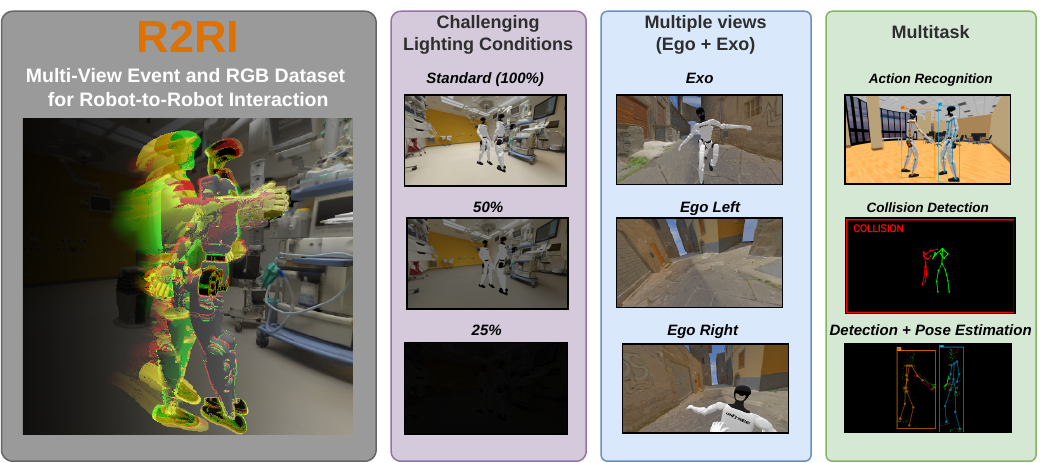}
    \caption{Overview of R2RI, a multi-view, multimodal dataset for Robot-to-Robot Interaction. R2RI provides synchronized RGB and event data from complementary exocentric and egocentric viewpoints, includes challenging illumination conditions, and supports multiple perception tasks, including robot detection, 2D/3D pose estimation, collision detection, and action recognition. \vspace{-12px}}
    \label{fig:placeholder}
\end{figure}

Despite the rapid evolution of robotic hardware and learning-based control systems~\cite{geravand2013human,kalaitzakis2021fiducial}, progress in data-driven interaction modeling remains constrained by the scarcity of large-scale, diverse, and publicly available interaction datasets. Existing data collections, as detailed in the following, often suffer from limited viewpoint diversity, limited robot types, narrow task definitions, or reliance on a single sensing modality. As a result, current models are bounded by scarce data availability~\cite{simoni20253d}, and as such, complex and complete analysis on their performances are limited. 
At the same time, acquiring real-world data is expensive, time-consuming, and difficult to scale due to strict safety constraints. Furthermore, the annotation process of real sensor streams is prone to noise and human error. To circumvent these limitations, synthetic data is a viable alternative as it provides a highly controllable, safe, and scalable environment to automatically generate massive amounts of perfectly annotated, multi-modal data.
Following these motivations, we present \textbf{Robot-to-Robot Interaction (R2RI)}, the first large-scale synthetic dataset specifically designed for the Robot-Robot Interaction (RRI) task. Several key properties distinguish R2RI: first, it consists of multiple models of recent commercial humanoid robots; second, it adopts a multiview capture strategy, providing both egocentric streams from each robot's onboard sensors and exocentric streams from static external cameras. Egocentric data reflects the information actually available to a robot during deployment, while exocentric data provides the global spatial context necessary for a complete understanding of the interaction. Third, the dataset is multimodal, offering synchronized RGB images and event camera data. 
Event cameras, with their microsecond temporal resolution and high dynamic range, are particularly well-suited to the fast, precise motions characteristic of robotic systems, yet remain underexplored in robot interaction datasets. Together, these modalities offer complementary representations that can support a diverse range of computer vision-based architectures.

Given that humanoid (or anthropomorphic) robots are increasingly resembling humans in both appearance and behavior~\cite{fink2012anthropomorphism}, the dataset is specifically constructed starting from a dataset containing human interactions (\textit{ie}, the recent INTER-X Dataset \cite{xu2024inter}). We propose a novel human-to-robot mapping algorithm, applicable to different robot models starting from their Unified Robot Description Format (URDF) model~\cite{quigley2009ros}. 
The dataset is created by rendering multiple controlled robots to simulate the same human interactions as the original dataset: this allowed for the automatic generation of a large number of frames with different robots, interaction type, background, and illumination conditions, with automatic annotation of each frame. The high frame rate (120 fps) enabled by the simulated environment allowed for the simultaneous rendering of RGB frames and event streams from different points of view.
In addition to dataset construction, we conduct a comprehensive comparative analysis of RGB and event-based approaches using state-of-the-art models for each task. This evaluation provides practical insights into modality-specific strengths and trade-offs, particularly regarding the modalities' complementary qualities.
The dataset and annotations are publicly released: we believe this resource will serve as a foundational benchmark for the emerging RRI task, and hope it will stimulate the development of robust, generalizable methods for autonomous agents operating in shared social environments.

In summary: \textbf{(i)} We introduce R2RI, the first large-scale multimodal dataset for robot-robot interaction, with heterogeneous humanoid platforms in egocentric and exocentric views. \textbf{(ii)} We propose a principled data collection and annotation methodology grounded in human interaction modeling. \textbf{(iii)} We provide comprehensive baseline evaluations across eight perception tasks using RGB and event modalities.

\section{Related Works}\label{sec2}

\begin{table}[t]
\centering
\caption{%
  Summary of the datasets available in literature proposed for Robot Manipulation (RM), Robot Pose Estimation (RPE), Human-Robot Interaction (HRI), and Robot-Robot Interaction (RRI). Further details are reported in Section~\ref{sec2}.
}
\label{tab:datasets}
\small
\resizebox{.95\textwidth}{!}{%
\begin{tabular}{lccccccl}
\toprule
  \textbf{Dataset} &  \textbf{Year} &  \textbf{\#Frames} &  \textbf{Images} &  \textbf{Multiview} &  \textbf{\#Robot} &  \textbf{Robot type} &  \textbf{Tasks} \\ 
\midrule

 {Open-X} \cite{o2024open}        & 2024 & --   & RGB-D        &                  & 22 & Arm      & RM       \\
 {Hum. Everyday} \cite{zhao2025humanoid}      & 2025 & --   & RGB-D, Lidar & \checkmark       & 2  & Humanoid & RM, HRI  \\
 {RoboInter} \cite{li2026robointer}           & 2026 & 86M  & RGB          & \checkmark       & 5  & Arm      & RM       \\

\midrule

 {CRAVES-lab} \cite{zuo2019craves}            & 2019 & 20k  & RGB          &                  & 1  & Arm      & RPE      \\ 
 {CRAVES-youtube} \cite{zuo2019craves}        & 2019 & 275  & RGB          &                  & 1  & Arm      & RPE      \\ 
 {Panda-3Cam} \cite{lee2020camera}            & 2020 & 17k  & RGB-D        &                  & 1  & Arm      & RPE      \\ 
 {Panda-Orb} \cite{lee2020camera}             & 2020 & 40k  & RGB-D        & \checkmark       & 1  & Arm      & RPE      \\ 
 {UNIBAS NAO} \cite{zampino2022fall}          & 2022 & 450  & RGB          &                  & 1  & Humanoid & RPE      \\
 {SimBa} \cite{simoni2022semi}                & 2022 & 20k  & RGB-D        &                  & 1  & Arm      & RPE      \\
 {DHRP} \cite{heo2024diverse}                 & 2024 & 16k  & RGB          &                  & 11 & Humanoid & RPE      \\
 {HRP} \cite{heo2024diverse}                  & 2024 & 1.5k & RGB          &                  & -- & Humanoid & RPE      \\
 {Unitree G1} \cite{young2025robots}          & 2025 & 8k   & RGB          &                  & 1  & Humanoid & RPE      \\

\midrule

 {Joker} \cite{devillers2015multimodal}       & 2015 & --   & RGB-D        &                  & 1  & Humanoid & HRI      \\ 
 {MHHRI} \cite{celiktutan2017multimodal}      & 2017 & --   & RGB-D        & \checkmark       & 1  & Humanoid & HRI      \\ 
 {UE-HRI} \cite{ben2017ue}                    & 2017 & --   & RGB-D, Lidar & \checkmark       & 1  & Humanoid & HRI      \\ 
 {MuMMER} \cite{canevet2020mummer}            & 2020 & 80k  & RGB-D        & \checkmark       & -- & Humanoid & HRI      \\ 
 {CHICO} \cite{sampieri2022pose}              & 2022 & 2M   & RGB          &                  & 1  & Arm      & HRI      \\

\midrule

 \textbf{R2RI (Ours)}                                       & 2026 & 6.5M & RGB-D, Event & \checkmark       & 4  & Humanoid & RPE, RRI \\ 

\bottomrule
\end{tabular}
}
\end{table}

\minisection{Robot-Robot Interaction Datasets.}
From a general perspective, to the best of our knowledge, there are no datasets in the literature explicitly designed to study the task of Robot-Robot Interaction. A summary of literature datasets is reported in Table~\ref{tab:datasets}.
Most of the existing datasets concerning the use of robots have been developed for the study of the Robotic Manipulation (RM) task~\cite{o2024open,zhao2025humanoid,li2026robointer}, and, despite their large size, they primarily focus on the use of non-humanoid robotic arms and end-effectors, manipulating static objects on a workspace surface.

Another major family of robotic datasets is devoted to Robot Pose Estimation (RPE), \textit{i.e.} the task of predicting the position of robot joints using only images~\cite{zuo2019craves,lee2020camera}. In these datasets, typically only a single robot is present, with annotations related to its pose, and no interactions are included. Finally, some datasets contain an extremely limited number of annotated frames or robot models~\cite{young2025robots,heo2024diverse}, which prevents the training and, more generally, the effective use of the most recent deep learning techniques available in the literature.
Other datasets in the literature have been specifically created for the Human–Robot Interaction (HRI) task. However, these datasets generally involve primarily verbal interaction, where one or more participants communicate with each other and, only at certain times, with the robot (Human-Human-Robot-Interaction, HHRI)~\cite{celiktutan2017multimodal}. Consequently, the datasets of this category do not provide annotations related to robot localization and pose~\cite{canevet2020mummer}.

\minisection{Robot Detection and Pose Estimation.}
Detecting the presence of robots in visual data has been approached from several directions~\cite{luo2017robot}. 
Early methods relied on engineered visual cues such as fiducial markers~\cite{kalaitzakis2021fiducial,garrido2014automatic}, specific sensors~\cite{hasegawa2010development}, or access to the robot’s proprietary software~\cite{geravand2013human} --  not always feasible --, achieving reliable detection only under controlled conditions and for known robot models. With the rise of deep Convolutional Neural Networks (CNN), detection approaches shifted toward learned representations, adapting general-purpose architectures such as Faster R-CNN~\cite{ren2015faster} and YOLO~\cite{khanam2024yolov11} to the robot detection task through fine-tuning on domain-specific data.

While robot pose estimation has recently gained significant research attention, most existing studies focus primarily on RGB images. 
From a general point of view, these solutions can be divided into two categories: i) hand–eye calibration methods and ii) rendering strategies. Hand–eye calibration methods typically rely on fiducial markers, such as ArUco patterns~\cite{garrido2014automatic}, usually attached to the robot’s end effector and observed by one or more cameras. However, these methods have an intrusive nature, due to the presence of markers on the robot, which may not always be feasible.
In contrast, rendering-based solutions~\cite{labbe2021single,noguchi2022watch} adopt a render-and-compare framework, in which the estimated pose is progressively optimized by repeatedly projecting a rendered model into the image and minimizing the discrepancy with the observed camera data.

Simba dataset~\cite{simoni2022semi} was one of the first datasets for 3D robot pose estimation~\cite{simoni2024dspdh}, including both RGB and depth information. DREAM~\cite{lee2020camera} introduced an encoder-decoder network to detect 2D keypoints of the robots and use PnP to recover extrinsics and then 3D pose; their authors also created a dataset which comprises both real and synthetic images of widely robot manipulators (\eg, Panda, Kuka, or Baxter robots), that is now one of the most affirmed baselines in the task. Despite the success of these methods, they solely focus on robot manipulators leaving out humanoid robots.

\minisection{Event-based Vision.}
Event cameras~\cite{delbruck2016neuromorophic,posch2014retinomorphic}, also known as neuromorphic cameras, have recently defined a new paradigm in vision sensing. Instead of synchronous frames, event cameras generate asynchronous streams of pixel-level activations (events), corresponding to brightness changes~\cite{lichtsteiner2008128}.
Neuromorphic sensors are effective because they generate sparse, motion-driven data, respond to logarithmic brightness changes with high dynamic range, perform well in low-light conditions, and capture events at microsecond-level temporal resolution, greatly reducing motion blur~\cite{event_ijprai,gallego2020event,magrini2025drone}.
Recently, event cameras have been used for a variety of applications (including tracking~\cite{seok2020robust,renner2019event,magrini2025fred}, object detection~\cite{liu2023motion,mitrokhin2018event,magrini2026pepr,magrini2024neuromorphic}, face analysis~\cite{shariff2023neuromorphic,becattini2025neuromorphic}), with a particular focus on robotics~\cite{mueggler2018event,mahlknecht2022exploring}, due to their low power consumption requirements.
Although prior research has explored human detection and pose estimation, to the best of our knowledge, this work represents the first application of event cameras to the tasks of robot to robot interaction.
\section{\method{} dataset}
\label{sec:method} 

\begin{figure}
    \centering
    \includegraphics[width=0.8\linewidth]{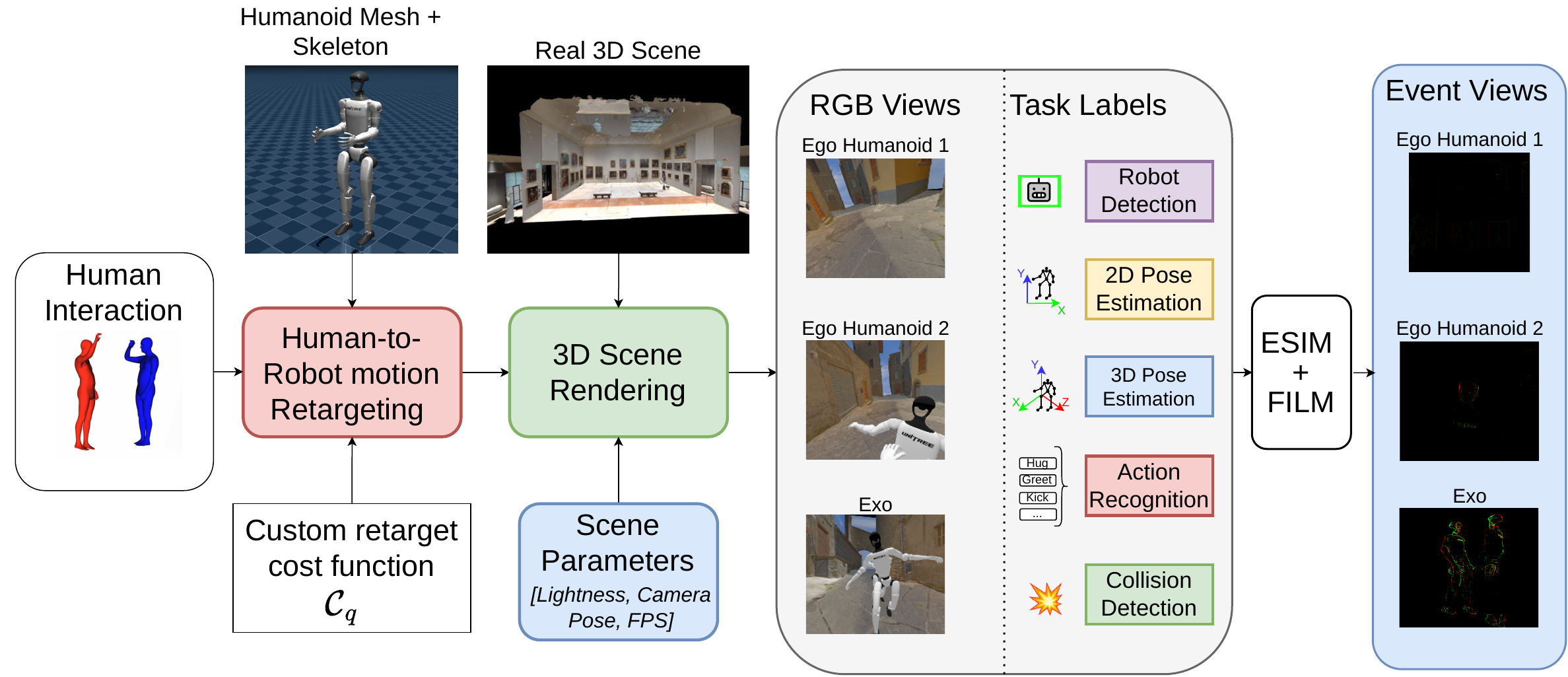}
    \caption{Overview of the {\method} dataset generation pipeline from synthetic robot meshes and real human interactions to the rendering of RGB and event frames.}
    \label{fig:schema}
\end{figure}

R2RI is a synthetic dataset containing videos of interactions between pairs of robots. Four humanoid robot models widely used in research were selected. The robots were placed in different 3D scenes, illuminated under different conditions and acquired by pairs of high-framerate stereo RGB cameras that allow the cogeneration of realistic event streams (as if captured by event cameras). One of the strengths of the dataset is the movement of the robots, which perform interactions by mimicking actions between pairs of real people. 
Four humanoid robots were chosen based on their popularity and availability of the URDF (Unified Robot Description Format) description: Boston Dynamics Atlas v5, Unitree G1, iCub, and  Aldebaran Robotics NAO. 
The description of the steps conducted for the dataset generation are reported in the next subsections, as well as depicted in Figure \ref{fig:schema}.

\minisection{Scene composition.}
Given a reference video collected from a human interaction dataset, a corresponding scene is created  in order to generate an output annotated video containing only robots instead of people. 
Thus, a robot model is selected in turn and associated to each person. 
The scene is created using \textit{pyrender} \cite{pyrender}, with ambient lighting configured to provide consistent illumination. Some scenes have a very dark illumination to simulate recordings at night.
For each time step $t$, the robots are animated using the Human to Robot motion retargeting described in the following subsection, which provides a robot pose relative to the root joint (\ie, the torso).

Absolute positions and orientations of each robot follow those of the corresponding person in the original video. A further ground alignment ensures that the robot's feet remain in contact with the floor, preserving physical plausibility. Specifically, the mesh is translated up or down (z-axis translation) until it intersects the floor plane.
The collision manager is updated with the transformed mesh positions at each frame to detect potential intersections between robots, reported in output as annotation of a contacting interaction. 
A global random rotation and translation are finally applied to the scene in order to introduce variability in the interaction, simulating different orientations relative to the camera or environment.
A series of 3D backgrounds representing both outdoor and indoor environments were used randomly to create variability in the scene.

\begin{figure}
    \centering
    \setlength{\tabcolsep}{2pt}
    \renewcommand{\arraystretch}{1.05}
    \begin{tabular}{cccc}
        \includegraphics[width=0.23\textwidth]{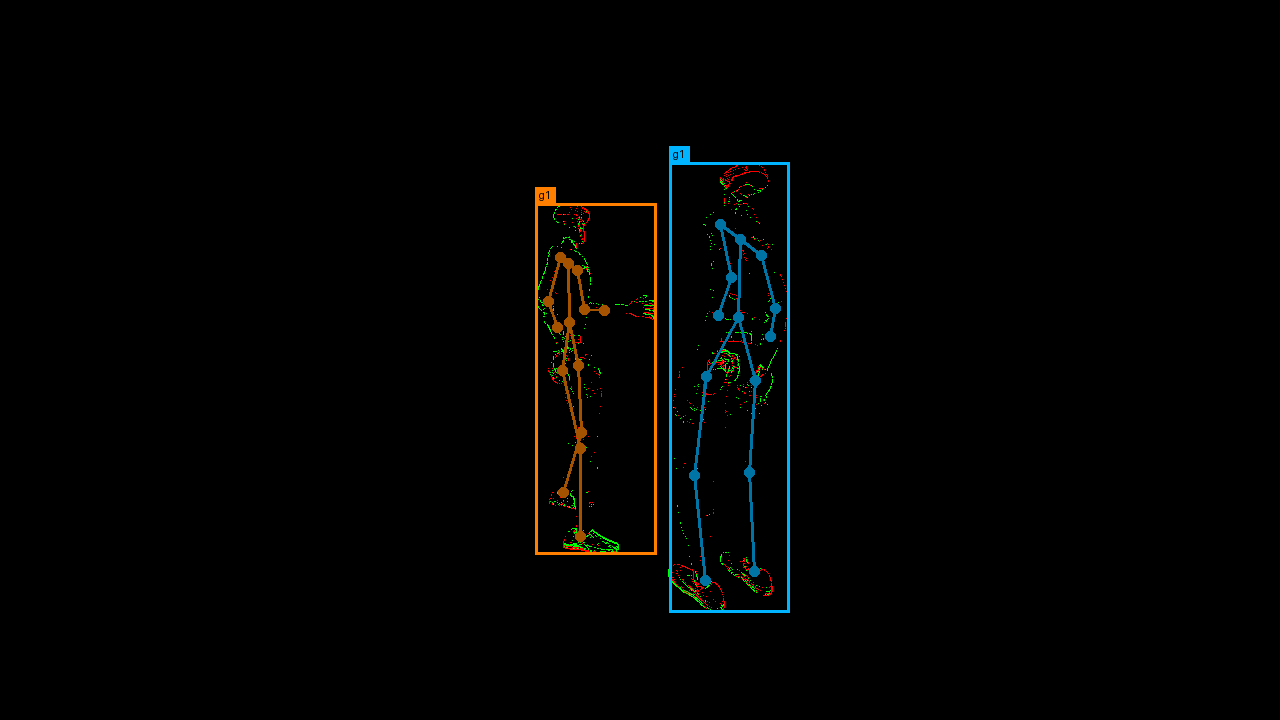} &
        \includegraphics[width=0.23\textwidth]{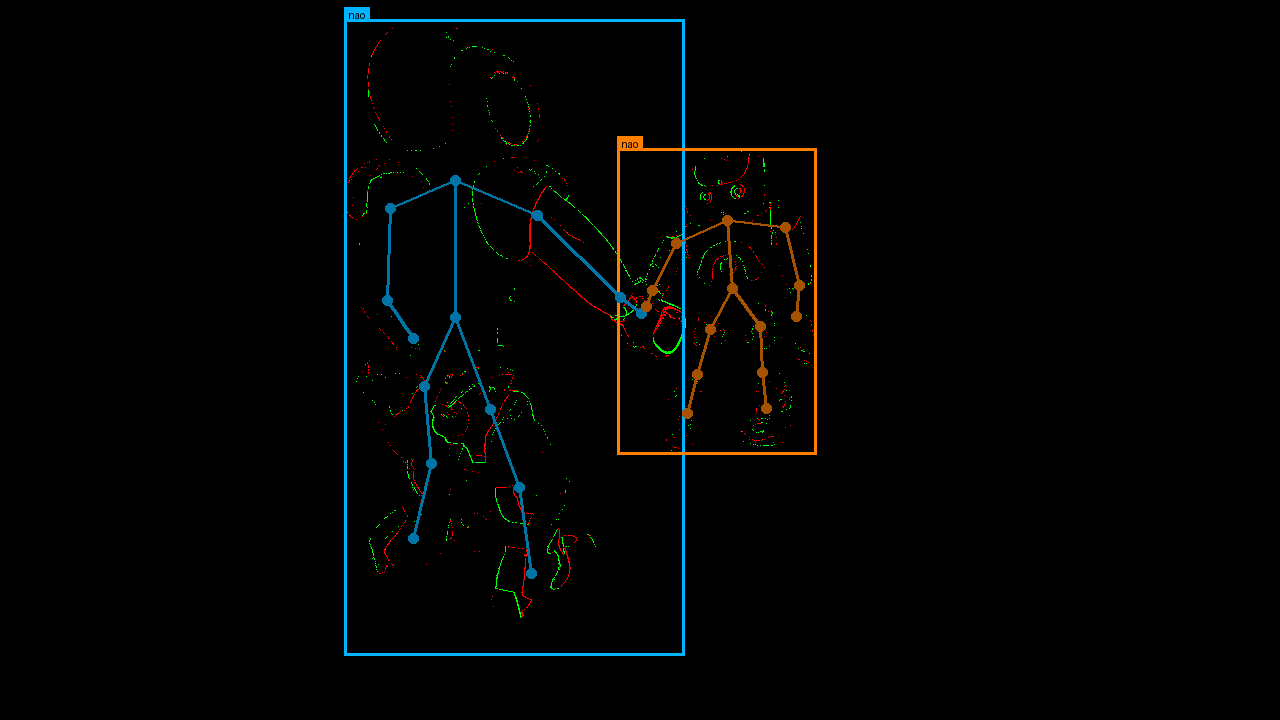} &
        \includegraphics[width=0.23\textwidth]{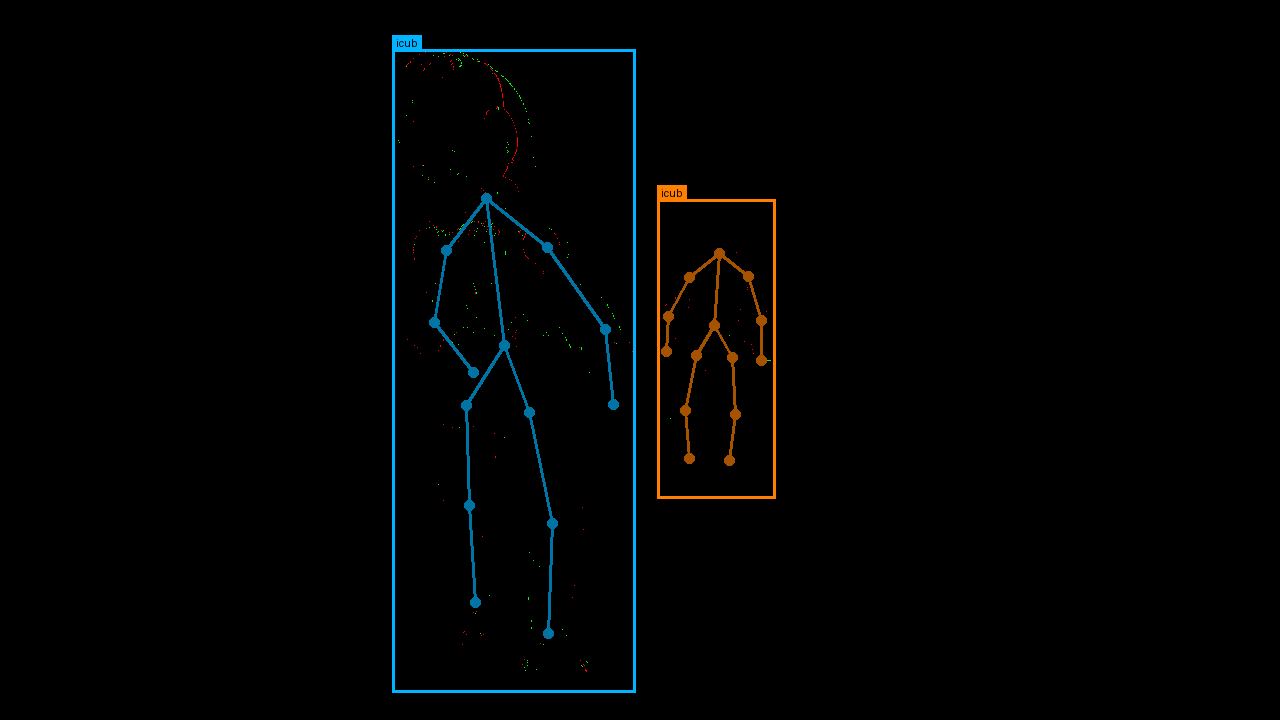} &
        \includegraphics[width=0.23\textwidth]{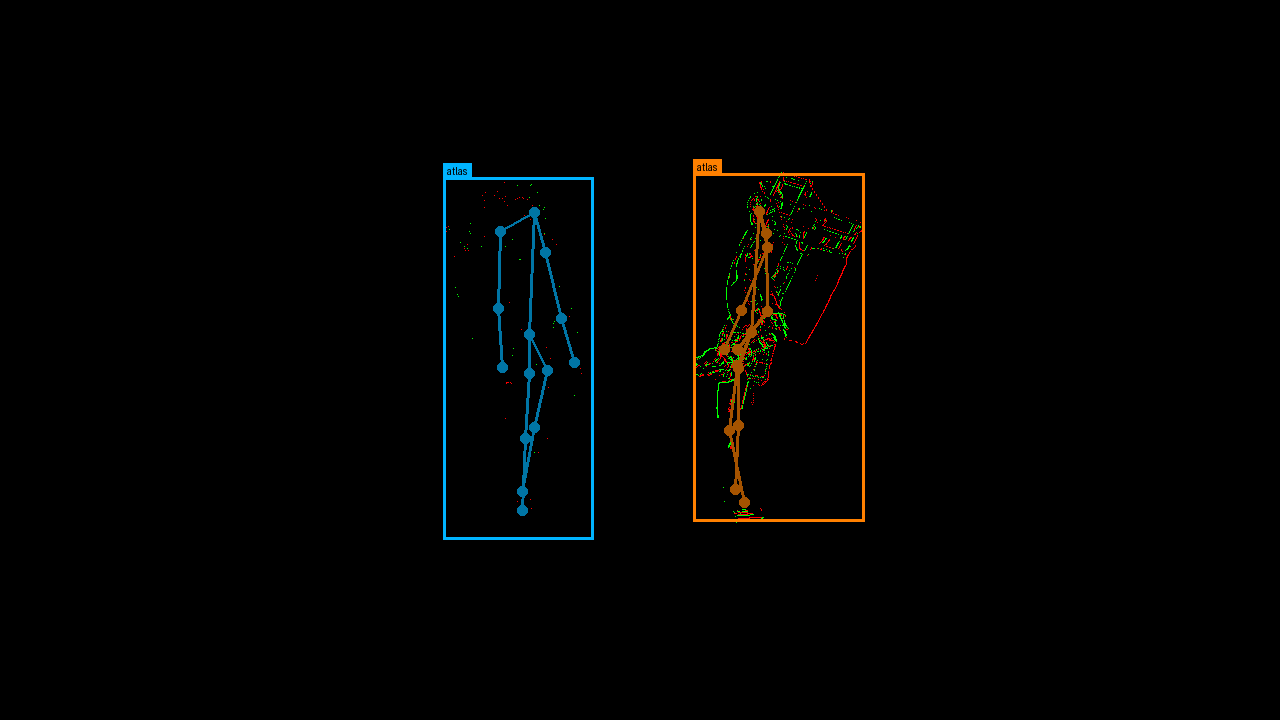} \\
        \includegraphics[width=0.23\textwidth]{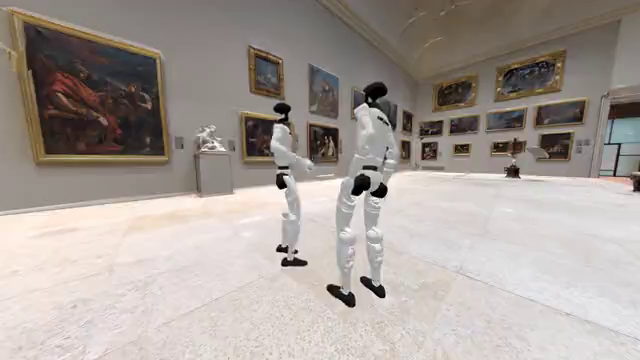} &
        \includegraphics[width=0.23\textwidth]{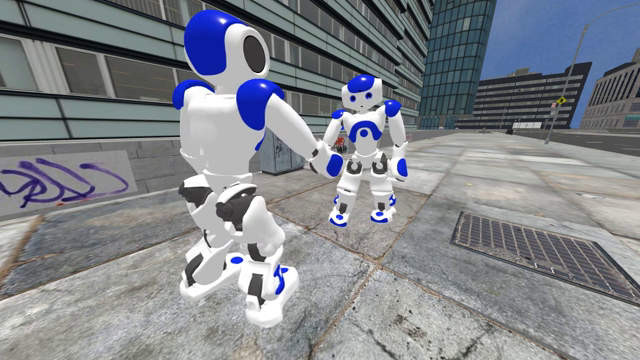} &
        \includegraphics[width=0.23\textwidth]{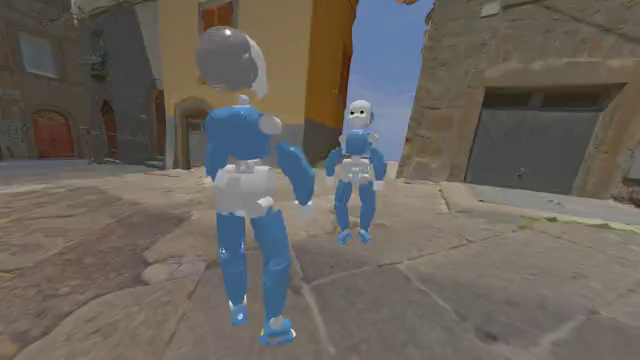} &
        \includegraphics[width=0.23\textwidth]{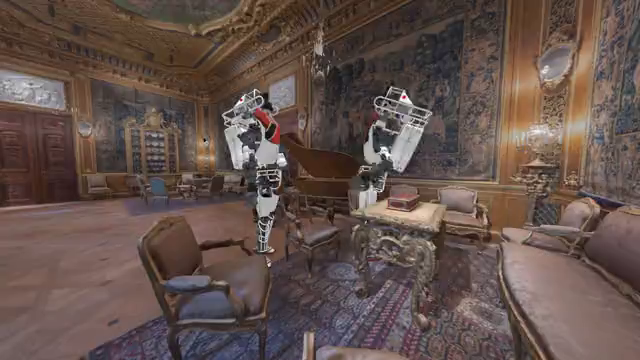} \\
    \end{tabular}
    \caption{Exo examples for both RGB and Event data. \method{} spans across a variety of environments and interactions, comprising both indoor and outdoor scenarios.}
    \label{fig:ego-exo-examples}
\end{figure}
\minisection{Human to Robot motion retargeting.} The dynamics of actions and interactions was extracted from an annotated dataset of human interactions. The motivations are multiple: to exploit the larger amount of data currently available on people compared to robots; to test the human-learning paradigm; and to automatically obtain an annotation of interactions with human-centric semantics.
For this purpose, we rely on sequences selected from a human interaction dataset. Among the others, we selected the INTER-X dataset \cite{xu2024inter}, which provides different classes of human interactions and a high frame rate setting. A subset of 20 interactions has been identified from the whole collection taking into account their portability to the robot domain. The list of selected interactions is reported in Appendix~\ref{sec:appendix_interactions}.

For each person, the sequence of human poses is represented using the SMPL-X model \cite{smplx}, thus providing a detailed parametric description of both body shape and pose. 
As a first step, we automatically converted the SMPL-X models into their corresponding skeleton-based poses using a $N_H=14$ joint model.

Let $J_i=\{\mathrm{x}_i,\mathrm{\theta}_i\}$ be the three spatial coordinates and the three orientation angles of each $i$-th joint obtained from the SMPL-X model.
Since current robots do not have the same shape and, more importantly, the same joints and degrees of freedom as humans, it is necessary to create a mapping between robot and human joints. Given a robot with $N_R$ joints (with $N_R$ depending on the specific robot selected and extracted from the corresponding URDF robot model), a mapping function $M(\cdot)$ is defined to provide the human joint corresponding (\ie, closer) to each robot joint. In addition, a global scale factor is introduced to make the robot and the person comparable. 

The final pose of the robot should be obtained by controlling its joint angles $\mathbf{q}$. In order to have the robot in the same position as the reference person, the joint angles $\mathbf{q}$ are obtained with an inverse kinematic framework using a constrained nonlinear optimization (SLSQP \cite{Joshy2024}) over a custom cost function $C(\mathbf{q})$ defined as follows:
\begin{equation}
C(\mathbf{q}) = \sum_{i=1}^{N_R} 
\left[ 
w_i \| \mathbf{FK}_x(\mathbf{q},i) - \mathrm{x}_{M(i)} \|^2 
+
\rho_i \left( \arccos (  \mathbf{FK}_\theta(\mathbf{q},i) \cdot  \mathrm{\theta}_{M(i)} ) \right)
\right].
\end{equation}

The cost function contains two terms, the first based on the distance of the mapped joints and the second based on the joint orientations. $w_i$ and $\rho_i$ are joint-wise weights, $\mathbf{FK}_x(\cdot,i)$ and $\mathbf{FK}_x(\cdot,i)$ provide the position and the orientation of the $i$-th joint of the robot given with the forward kinematics algorithm. $\mathrm{x}_{M(i)}$ and $\mathrm{\theta}_{M(i)}$ are the position and orientation of the human joint assigned to the $i$-th robot joint (and retrieved thanks to the mapping function $M(\cdot)$).

The optimization is performed for each frame and for each robot in the scene starting from the current pose $\mathbf{q}_0$ as reference. The output of the optimizer is the final pose of the robot expressed in motor angles
    $\mathbf{q}^* = \arg\min_{\mathbf{q}} C(\mathbf{q})$.

\minisection{Rendering.}
The rendering of each timestep starts by loading precomputed robot poses and scaling factors from as previously described.  
Each robot mesh is added to the scene as a distinct node, associated with its corresponding reference frame to maintain correct spatial alignment.

Several virtual cameras are placed in the scene in order to generate a multi-view and multi-modal dataset. Specifically, an RGB camera pair (\ie, a stereo camera) is fixed and placed to have an exocentric view of the scene. Similary, two RGB cameras follows the robot head to simulate egocentric views.  
The exocentric camera is positioned dynamically to frame the interaction between the robots, centering on the midpoint between the two robots while maintaining a horizontal orientation and applying vertical and lateral offsets to ensure optimal coverage. Camera extrinsic parameters are exported for each frame relative to the world coordinate system.

\begin{table}[t]
\centering
\caption{\textbf{R2RI dataset statistics.}
R2RI provides synchronized RGB and event streams from exo and ego viewpoints at 120\,fps.}
\label{tab:dataset_stats}
\footnotesize
\setlength{\tabcolsep}{5pt}
\renewcommand{\arraystretch}{0.95}
\begin{tabular}{l c}
\toprule
\multicolumn{2}{c}{\textbf{Dataset overview}} \\
\midrule
\textbf{Attribute} & \textbf{Value} \\
\midrule
Sequences & $\approx$5000 \\
Interaction classes & 20 \\
Robot platforms & Atlas, G1, iCub, NAO \\
Viewpoints & Exocentric, Egocentric (2 egos) \\
Modalities & RGB, Depth, Event \\
Lighting conditions & Standard (100\%), 50\%, 25\%, 5\% \\
Frame rate & 120\,fps \\
Clip duration & 1.8--23.5\,s; mean 5.8\,s \\
Frames & 6.5 million \\
\bottomrule
\end{tabular}
\end{table}

During the rendering, 3D positions of each joints are saved, using the global reference frame and each robot/camera reference frame. In addition, their 2D projection on each camera image plane are generated and saved.
The rendering frame rate follows the original one (\ie, 120 fps).

\minisection{Event Simulation.}
To create realistic event camera data, we use the ESIM simulator~\cite{Rebecq18corl} in conjunction with FILM frame interpolation~\cite{reda2022film}, following the video-to-event pipeline of~\cite{hu2021v2e}. In order to facilitate the event simulation and final quality, we also generated the original RGB robot frames at 120 FPS, leading to a smoother frame interpolation.

\section{Experimental Settings and Baselines}
\label{sec:exp_settings}
The dataset supports perception models for detecting the presence, pose, and motion of surrounding and interacting robots (Fig.~\ref{fig:ego-exo-examples}). We define eight tasks covering common interaction scenarios. Since the dataset includes egocentric and exocentric views, partial visibility, self-occlusions, and mutual occlusions are key challenges.
\minisection{Robot Detection.} Given a single RGB or event frame, the task is to detect each robot and its 2D bounding box. Performance is measured with standard detection metrics~\cite{LinMBHPRDZ14}: $mAP_{50}$ and $mAP_{50:95}$.
\minisection{2D Robot Pose Estimation.} The target joints depend on the robot model. Following COCO~\cite{LinMBHPRDZ14}, performance is measured using $mAP_{50}$ and $mAP_{50:95}$.
\minisection{3D Robot Pose Estimation.} The task is to estimate the complete 3D robot pose as 3D joint coordinates. Synthetic annotations provide precise internal joint positions, including occluded joints. Predictions are evaluated using MPJPE and P-MPJPE~\cite{Rhodin2018}.
\minisection{2D Pose Forecasting.} Given 2D keypoints of both robots over $T_\text{obs}=15$ frames, the task is to predict the next $T_\text{pred} \in \{15, 30\}$ frames. Sequences are subsampled from 120\,fps to 30\,fps. Performance is measured by ADE and FDE in pixels, averaged over predicted time steps and joints.
\minisection{3D Pose Forecasting.} Given $T_\text{obs}=15$ observed 3D joint positions for both robots, the task is to predict the next $T_\text{pred} \in \{15, 30\}$ frames. Performance is reported as ADE and FDE, both in millimetres.
\minisection{Robot Collision Detection.} Given a video clip, the task is to predict frame-level physical contact between the two robots. Labels are generated by the physics-engine collision manager. We evaluate a privileged pose-based setting using ground-truth 3D joints and a visual setting using RGB or event clips. Performance is reported using F1, balanced accuracy, and AUROC.
\minisection{Robot-to-Robot Action Recognition.} Given a clip of two interacting robots, the task is to classify the interaction among 20 categories derived from human social behaviours. Evaluation is performed independently on RGB and event-camera inputs.
\minisection{Sim-to-Real.} We evaluate zero-shot transfer from synthetic training data to real-world robot videos, without fine-tuning. Transfer is measured with the same metrics as the corresponding source task. We choose in particular the DHRP~\cite{heo2024diverse} dataset as it presents a decisive variety of humanoids, including some extremely different in shapes and size with respect to the ones present in R2RI.

To establish replicable benchmarks, we define four experimental settings with fixed train, validation, and test splits, released with the dataset.
\minisection{Standard setting.} The standard setting uses R2RI-Full under standard illumination, enabling fair RGB/event comparison. Splits include all robots, backgrounds, and interactions.
\minisection{Challenging Lighting Conditions.} We evaluate performance under reduced illumination using the {\method}-Night subset, where lightness is reduced to 50\%, 25\% and 5\% of the standard setting (referred to as Half, Quarter and Min.\ brightness in the tables), reporting RGB and event results separately.
\minisection{Humanoid Generalization.} We evaluate cross-platform generalization with Leave-One-domain-Out testing on {\method}-Full, Ego1, and Night splits. Models are trained on all but one humanoid robot and tested on the held-out platform.
\minisection{Viewpoint Generalization.} This setting evaluates generalization between exocentric and egocentric viewpoints. Models trained on external cameras are tested on onboard views, and vice versa, introducing changes in perspective, scale, and occlusion.

We trained several baselines for each task on our data in each modality. For the task of 2D Robot Detection we used two-stage~\cite{ren2015faster} and single stage~\cite{khanam2024yolov11, sapkota2025yolo26} convolutional detectors and a selection of more recent transformer based detectors~\cite{carion2020end,zhu2020deformable, zhang2022dino, zhao2024detrs}. 
Specifically, we evaluated Faster R-CNN~\cite{ren2015faster}, YOLOv11~\cite{khanam2024yolov11}, YOLO26~\cite{sapkota2025yolo26}, DETR~\cite{carion2020end}, Deformable DETR~\cite{zhu2020deformable}, DINO~\cite{zhang2022dino}, and RT-DETR~\cite{zhao2024detrs}. For both YOLO families, we evaluate multiple checkpoints to better assess the efficiency/accuracy trade-off.
For the task of 2D Robot Pose Estimation, we used a selection of convolutional pose estimators~\cite{jiang2023rtmpose,yang2023effective,sun2019deep, khanam2024yolov11, sapkota2025yolo26}, including RTMPose~\cite{jiang2023rtmpose}, its distillation enhanced evolution~\cite{yang2023effective}, HRNet~\cite{sun2019deep}, and pose detectors based on the YOLO architecture~\cite{khanam2024yolov11, sapkota2025yolo26}. 
We utilize the implementations provided by the \texttt{mmdet} and \texttt{ultralytics} Python libraries for the object detection models, and \texttt{mmpose} together with \texttt{ultralytics} for the 2D pose estimation methods.
All models were trained for 20 epochs.
For the task of 3D Robot Pose Estimation, we explored monocular video-based models employing three different state-of-the-art pose estimation architectures: PoseFormerV2~\cite{Zhao_2023_CVPR}, MotionBERT~\cite{motionbert2022}, and MotionAGFormer~\cite{mehraban2024motionagformer}. These models were chosen for their transformer-based approach, which is able to exploit temporal information across frames to improve 3D pose reconstruction. During training, only 2D ground-truth keypoints were used as supervision.

\begin{table}[t]
    \centering
\begin{minipage}[t]{0.49\linewidth}\centering
\caption{2D robot detection results on the exocentric test split.}
\label{tab:detection_results}
\resizebox{1\linewidth}{!}{
\begin{tabular}{lcccc}
\toprule
\multirow{2}{*}{\textbf{Model}} 
& \multicolumn{2}{c}{\textbf{Event}} 
& \multicolumn{2}{c}{\textbf{RGB}} \\
\cmidrule(lr){2-3}\cmidrule(lr){4-5}
& mAP$_{50}$\,$\uparrow$ & mAP$_{50:95}$\,$\uparrow$ 
& mAP$_{50}$\,$\uparrow$ & mAP$_{50:95}$\,$\uparrow$ \\
\midrule
Faster R-CNN~\cite{ren2015faster}        & 92.0 & 66.8 & 93.9 & 76.7 \\
Deformable-DETR~\cite{zhu2020deformable} & 97.8 & 72.3 & 96.1 & 81.2 \\
RT-DETR~\cite{zhao2024detrs}             & 97.3 & 82.1 & 94.4 & \underline{88.5} \\
DINO~\cite{zhang2022dino}                & 97.4 & 74.6 & 97.3 & \underline{84.2} \\
DETR~\cite{carion2020end}                & 96.4 & 67.0 & 93.2 & 66.1 \\
YOLO11l~\cite{khanam2024yolov11}         & \underline{98.4} & \underline{86.8} & \underline{97.8} & \textbf{94.2} \\
YOLO26l~\cite{sapkota2025yolo26}         & \textbf{98.8} & \textbf{89.1} & \textbf{98.1} & \textbf{94.2} \\
\bottomrule
\end{tabular}%
}
\end{minipage} \hfill
\begin{minipage}[t]{0.49\linewidth}\centering
\caption{2D Pose Estimation results on the exocentric test split.}
\label{tab:pose_results_full}
\resizebox{1\linewidth}{!}{%
\begin{tabular}{lcccc}
\toprule
\multirow{2}{*}{\textbf{Model}} & \multicolumn{2}{c}{\textbf{Event}} 
& \multicolumn{2}{c}{\textbf{RGB}} \\
\cmidrule(lr){2-5}
& mAP$_{50}$\,$\uparrow$ & mAP$_{50:95}$\,$\uparrow$ 
& mAP$_{50}$\,$\uparrow$ & mAP$_{50:95}$\,$\uparrow$ \\
\midrule
RTMPose~\cite{jiang2023rtmpose} & 89.9 & 68.8 & 97.0 & 93.9 \\
DWPose~\cite{yang2023effective} & 89.9 & 69.3 & 97.0 & 93.6 \\
HRNet~\cite{sun2019deep}        & 83.7 & 62.0 & 97.0 & 92.5 \\
YOLO11l-pose~\cite{khanam2024yolov11} & 94.2 & 78.0 & 97.2 & 91.3 \\
YOLO26l-pose~\cite{sapkota2025yolo26} & \textbf{95.7} & \textbf{81.3} & \textbf{97.3} & \textbf{91.9} \\
\bottomrule
\end{tabular}
}
\end{minipage} \hfill
\end{table}

\section{Experimental Results}
\label{sec:results}
\minisection{2D Robot Detection.}
In Table~\ref{tab:detection_results}, we show the performances of the detection models for both RGB and event frames using the exocentric point of view. 
YOLO26l~\cite{sapkota2025yolo26} demonstrated to be the best in both metrics for both event and RGB modalities.

\begin{table}[t]
    \centering
\begin{minipage}[t]{0.39\linewidth}\centering
\caption{3D Pose Estimation Results, tested on models trained in RGB modality.}
\label{tab:3d_pose}
\setlength{\tabcolsep}{6pt}
\resizebox{\linewidth}{!}{
\begin{tabular}{lcc}
\toprule
\textbf{Model} & \textbf{MPJPE} $\downarrow$ & \textbf{P-MPJPE}$\downarrow$ \\
\midrule
PoseFormerV2~\cite{Zhao_2023_CVPR} & 55.9 & 35.8  \\
MotionAGFormer~\cite{mehraban2024motionagformer}  & \underline{29.4} & \underline{23.2}  \\
MotionBERT~\cite{motionbert2022}  & \textbf{25.6} & \textbf{20.5} \\
\bottomrule
\end{tabular}
}
\end{minipage} \hfill
\begin{minipage}[t]{0.58\linewidth}\centering
\caption{Detection and pose estimation results in the ``Challenging Lighting Conditions'' setting.}
\label{tab:night_split_combined}
\resizebox{1\linewidth}{!}{%
\begin{tabular}{l c c cccc}
\toprule
\textbf{Task} & \textbf{Mod.} & \textbf{Model} 
& \multicolumn{2}{c}{\textbf{Half brightness}} 
& \multicolumn{2}{c}{\textbf{Min. brightness}} \\
\cmidrule(lr){4-5} \cmidrule(lr){6-7}
& & 
& mAP$_{50}$\,$\uparrow$ & mAP$_{50:95}$\,$\uparrow$ 
& mAP$_{50}$\,$\uparrow$ & mAP$_{50:95}$\,$\uparrow$ \\
\midrule

\multirow{2}{*}{Det.}
& Event & \multirow{2}{*}{YOLO26l~\cite{sapkota2025yolo26}}
& 87.7 & 65.3 & \textbf{88.1} & \textbf{66.0} \\

& RGB & 
& \textbf{98.3} & \textbf{94.2} & 0.7 & 0.3 \\

\midrule

\multirow{2}{*}{Pose}
& Event & \multirow{2}{*}{YOLO26l-pose~\cite{sapkota2025yolo26}}~~~ 
& 88.6 & 67.8 & \textbf{86.7} & \textbf{66.5} \\

& RGB &  
& \textbf{97.9} & \textbf{90.4} & 1.9 & 0.5 \\

\bottomrule
\end{tabular}
}
\end{minipage}
\hfill
\end{table}

\minisection{2D and 3D Pose Estimation.}
Similarly to 2D detection, we employ a roster of state-of-the-art models for benchmarking the exocentric 2D Pose Estimation task. 
In Table~\ref{tab:pose_results_full}, the results for both the event and RGB domains are reported. Also in this case, YOLO26l-pose~\cite{sapkota2025yolo26} showed superior performance.
The detection and pose estimation results from the egocentric point of view are reported further on, for simplicity, in Tables~\ref{tab:det_eval} and~\ref{tab:pose_eval}.
Table~\ref{tab:3d_pose} reports 3D Pose Estimation results using MPJPE and P-MPJPE. All models were evaluated with the same input setup (81 frames per sequence, stride 27), ensuring a fair comparison. MotionBERT~\cite{motionbert2022} outperforms both competitors, indicating that its temporal modeling is more effective at capturing motion patterns over long sequences.


\minisection{Challenging Lighting Conditions} Table~\ref{tab:night_split_combined} reports the best performing methods for the detection (YOLO26l) and pose estimation (YOLO26l-pose) tasks under sever lightness degradation. While RGB models performances drop consistently in the almost total absence of light, performance remains consistent for the event models, showing the advantage of having that modality for both tasks.
Complete results for all detection and 2D pose estimation models at every illumination level are reported in Appendix~\ref{sec:appendix_challenge}.

\begin{table}[t]
\centering
\footnotesize
\caption{Results for the ``Humanoid Generalization'' setting, using a leave-one-domain-out protocol. 
}
\label{tab:method_dg_all}
\resizebox{\textwidth}{!}{
\setlength{\tabcolsep}{4pt}
\begin{tabular}{l c ccccccccc}
\toprule
\multirow{2}{*}{\textbf{Task}} & \multirow{2}{*}{\textbf{Model}} & \multirow{2}{*}{\textbf{Mod}.} 
& \multicolumn{2}{c}{\textbf{Atlas Out}}
& \multicolumn{2}{c}{\textbf{G1 Out}}
& \multicolumn{2}{c}{\textbf{iCub Out}}
& \multicolumn{2}{c}{\textbf{NAO Out}} \\
\cmidrule(lr){4-5} \cmidrule(lr){6-7} \cmidrule(lr){8-9}  \cmidrule(lr){10-11}
&
& 
& mAP$_{50}$\,$\uparrow$ & mAP$_{50:95}$\,$\uparrow$
& mAP$_{50}$\,$\uparrow$ & mAP$_{50:95}$\,$\uparrow$
& mAP$_{50}$\,$\uparrow$ & mAP$_{50:95}$ \,$\uparrow$
& mAP$_{50}$\,$\uparrow$ & mAP$_{50:95}$\,$\uparrow$\\
\midrule
\multirow{2}{*}{Det.}
&
\multirow{2}{*}{YOLO26l~\cite{sapkota2025yolo26}}
& Event 
& \textbf{91.5} & \textbf{71.6}
& \textbf{93.4} & \textbf{69.1}
& \textbf{92.1} & \textbf{67.9}
& 91.6 & \textbf{67.5} \\
& & RGB 
& 50.6 & 21.1
& 76.5 & 38.5
& 79.0 & 39.0
& \textbf{93.8} & 66.6 \\
\midrule
\multirow{2}{*}{Pose}
&
\multirow{2}{*}{YOLO26l-pose~\cite{sapkota2025yolo26}}
& Event 
& \textbf{26.7} & \textbf{6.7}
& \textbf{53.5} & 1\textbf{6.3}
& \textbf{60.9} & \textbf{25.0}
& \textbf{56.6} & \textbf{21.1} \\
& & RGB 
& 25.6 & 4.6
& 36.1 & 11.7
& 48.9 & 8.2
& 22.1 & 7.7 \\

\bottomrule
\end{tabular}
}
\end{table}
\begin{table}[t]
    \centering
\begin{minipage}[t]{0.48\linewidth}\centering
\caption{Detection evaluation across domains (``Viewpoint Generalization'' setting).}
\label{tab:det_eval}
\resizebox{\textwidth}{!}{
\setlength{\tabcolsep}{4pt}
\begin{tabular}{l c c c c c c c}
\toprule
\textbf{Model} & \textbf{Mod.} 
& \multicolumn{2}{c}{\textbf{On-domain (Ego2Ego)}} 
& \multicolumn{2}{c}{\textbf{Cross-domain (Exo2Ego)}} 
& \multicolumn{2}{c}{\textbf{Cross-domain (Ego2Exo)}} \\
\cmidrule(lr){3-4} \cmidrule(lr){5-6} \cmidrule(lr){7-8}
& 
& mAP$_{50}$\,$\uparrow$ & mAP$_{50:95}$\,$\uparrow$
& mAP$_{50}$\,$\uparrow$ & mAP$_{50:95}$\,$\uparrow$ 
& mAP$_{50}$\,$\uparrow$ & mAP$_{50:95}$\,$\uparrow$ \\
\midrule

\multirow{2}{*}{Def-DETR\cite{zhu2020deformable}}
& Event & 95.5 & 79.9 & 0.1 & 0.0 & 33.0 & \textbf{16.3} \\
& RGB   & \textbf{97.7} & \textbf{89.3} & \textbf{62.6} & \textbf{43.6} & \textbf{33.8} & 12.9 \\
\midrule

\multirow{2}{*}{DINO\cite{zhang2022dino}}
& Event & 95.2 & 84.0 & 0.0 & 0.0 & \textbf{39.6} & \textbf{19.1} \\
& RGB   & \textbf{98.4} & \textbf{95.3} & \textbf{88.4} & \textbf{79.2} & 36.2 & 15.3 \\
\midrule

\multirow{2}{*}{YOLO26l\cite{sapkota2025yolo26}}
& Event & 97.6 & 91.1 & 1.9 & 1.1 & \textbf{65.2} & \textbf{49.7} \\
& RGB   & \textbf{98.4} & \textbf{95.4} & \textbf{83.5} & \textbf{76.7} & 62.7 & 36.5 \\

\bottomrule
\end{tabular}
}
\end{minipage} \hfill
\begin{minipage}[t]{0.49\linewidth}\centering

\caption{Pose evaluation across domains (``Viewpoint Generalization'' setting).}
\label{tab:pose_eval}
\resizebox{\textwidth}{!}{
\setlength{\tabcolsep}{4pt}
\begin{tabular}{l c c c c c c c}
\toprule
\textbf{Model} & \textbf{Mod.}
& \multicolumn{2}{c}{\textbf{On-domain (Ego2Ego)}}
& \multicolumn{2}{c}{\textbf{Cross-domain (Exo2Ego)}}
& \multicolumn{2}{c}{\textbf{Cross-domain (Ego2Exo)}} \\
\cmidrule(lr){3-4} \cmidrule(lr){5-6} \cmidrule(lr){7-8}
&
& mAP$_{50}$\,$\uparrow$ & mAP$_{50:95}$\,$\uparrow$
& mAP$_{50}$\,$\uparrow$ & mAP$_{50:95}$\,$\uparrow$
& mAP$_{50}$\,$\uparrow$ & mAP$_{50:95}$\,$\uparrow$ \\
\midrule

\multirow{2}{*}{RTMPose\cite{jiang2023rtmpose}}
& Event & 81.5 & 60.0 & 11.6 & 5.4 & 44.0 & 20.9 \\
& RGB   & \textbf{93.9} & \textbf{88.1} & \textbf{61.2} & \textbf{49.7} & \textbf{79.5} & \textbf{62.6} \\
\midrule

\multirow{2}{*}{DWPose\cite{yang2023effective}}
& Event & 80.4 & 59.2 & 11.7 & 5.3 & 42.8 & 20.0 \\
& RGB   & \textbf{92.9} & \textbf{87.3} & \textbf{60.5} & \textbf{48.9} & \textbf{81.1} & \textbf{65.2} \\
\midrule

\multirow{2}{*}{YOLO26l-pose\cite{sapkota2025yolo26}}
& Event & 94.7 & \textbf{83.7} & 0.6 & 0.3 & \textbf{60.6} & \textbf{38.9} \\
& RGB   & \textbf{94.8} & 80.7 & \textbf{75.4} & \textbf{61.6} & 39.7 & 21.2 \\

\bottomrule
\end{tabular}
}
\end{minipage} \hfill
\end{table}
\begin{table}[t]
\centering
\caption{2D and 3D pose forecasting. 2D units: pixels; 3D units: millimetres.}
\label{tab:forecasting}
\resizebox{.8\textwidth}{!}{
\begin{tabular}{l rr rr rr rr}
\toprule
 & \multicolumn{4}{c}{2D pose} & \multicolumn{4}{c}{3D pose} \\
\cmidrule(lr){2-5}\cmidrule(lr){6-9}
 & \multicolumn{2}{c}{Horizon $h=15$} & \multicolumn{2}{c}{Horizon $h=30$}
 & \multicolumn{2}{c}{Horizon $h=15$} & \multicolumn{2}{c}{Horizon $h=30$} \\
\cmidrule(lr){2-3}\cmidrule(lr){4-5}\cmidrule(lr){6-7}\cmidrule(lr){8-9}
Model & ADE\,$\downarrow$ & FDE\,$\downarrow$ & ADE\,$\downarrow$ & FDE\,$\downarrow$ & ADE\,$\downarrow$ & FDE\,$\downarrow$ & ADE\,$\downarrow$ & FDE\,$\downarrow$ \\
\midrule
Linear Velocity & 25.2 & 50.2 & 41.0 & 82.0 & 47.3 & 99.7  & \textbf{100.4} & 208.6 \\
MLP (512-256)   & 58.8 & 68.4 & 61.2 & 77.3 & 134.8 & 148.5 & 159.1 & 191.2 \\
TCN~\cite{lea2017temporal}             & 47.7 & 63.9 & 57.6 & 75.2 & 77.4 & \textbf{103.2} & 117.2 & 173.9 \\
GRU~\cite{dey2017gate}     & \textbf{45.0} & \textbf{56.3} & 53.7 & 76.0 & \textbf{89.0} & 112.4 & 130.9 & 182.4 \\
Transformer     & \textbf{45.0} & \textbf{56.3} & \textbf{49.5} & \textbf{67.7} & 98.9 & 117.5 & 126.1 & \textbf{165.7} \\
\bottomrule
\end{tabular}
}
\end{table}

\minisection{Humanoid Generalization.}
In Table~\ref{tab:method_dg_all} we report the result about the Humanoid Generalization setting. Event-based inputs consistently outperform RGB across domains in object detection, especially in the mAP$_{50:95}$ metric, demonstrating more generalization capability. RGB performs comparably only on domains that are visually similar to the train set (\eg, NAO). Both modalities struggle with the pose estimation task. Overall, event modality shows greater robustness on domain shifts.

\minisection{Viewpoint Generalization.} the results are reported in Tables~\ref{tab:det_eval} and~\ref{tab:pose_eval}. When training on the exocentric train set and testing on the egocentric set (Exo2Ego), we notice how RGB outperforms event in both tasks. RGB inputs likely benefit from strong visual cues (colors, textures, backgrounds) that are consistent with exo training, which transfer well to ego frames. Event data, being sparse and edge-based, may lose some of this context, leading to lower performance. When training on the ego training set and testing on the exo test set, performances are similar. RGB does not gain a clear advantage in this cross-evaluation, probably because exo frames are visually different from the ego training set, so their domain-specific features transfer less effectively. 

\minisection{Pose Forecasting.} Table~\ref{tab:forecasting} reports 2D and 3D pose forecasting results for two prediction horizons ($h\!=\!15$ and $h\!=\!30$). As expected, errors increase substantially when doubling the horizon, highlighting the difficulty of long-term forecasting under viewpoint changes and occlusions. For 2D forecasting, sequence models clearly outperform the simple baselines: GRU and Transformer achieve the best short-horizon performance (ADE/FDE), while the Transformer is the most accurate at $h\!=\!30$, suggesting that attention helps preserve temporal consistency over longer futures. In 3D, the gap between models is smaller and the constant-velocity baseline is surprisingly competitive, achieving the best ADE at $h\!=\!30$, which indicates that many interactions exhibit locally smooth joint trajectories; nevertheless, learned temporal models (TCN/GRU/Transformer) consistently improve endpoint accuracy (FDE), particularly at longer horizons.

\minisection{Collision Detection.} We report collision detection results in Table~\ref{tab:collision_results}. In the privileged setting (GT 3D pose), simple pose-based classifiers are effective: both Random Forest and a shallow MLP achieve very high F1 and AUROC, confirming that collisions are largely linearly/separably predictable from short-term joint geometry and relative motion cues. In the visual setting, performance drops due to occlusions and appearance variability. However, event frames yield higher balanced accuracy and AUROC than RGB for the same ResNet-50 backbone, suggesting that motion-centric event representations provide a cleaner signal for physical contact under challenging lighting and clutter.

\begin{table}[t]
\centering
\footnotesize
\setlength{\tabcolsep}{3pt}
\caption{Collision detection(a) and Action recognition(b) results.}
\label{tab:benchmark_results}

\begin{subtable}[t]{0.5\textwidth}
\centering
\caption{Collision detection. Top scoring results are extracted using 3D Poses ground-thruths (GT), while frame based approaches still obtain good results.}
\label{tab:collision_results}
\begin{tabular}{llccc}
\toprule
Model & Input & F1\,$\uparrow$ & BalAcc\,$\uparrow$ & AUROC\,$\uparrow$ \\
\midrule
Random Forest & GT 3D pose & \textbf{0.949} & \textbf{0.841} & \textbf{0.977} \\
MLP (256-128) & GT 3D pose & \underline{0.947} & \underline{0.867} & \underline{0.947} \\
\midrule
ResNet-50~\cite{he2016deep}     & RGB frames   & 0.771 & 0.584 & 0.654 \\
ResNet-50~\cite{he2016deep}     & Event frames & 0.774 & 0.737 & 0.808 \\
\bottomrule
\end{tabular}
\end{subtable}
\hfill
\begin{subtable}[t]{0.45\textwidth}
\centering
\caption{Action recognition (20 classes).Resnet-50~\cite{he2016deep} uses a mean-pool (MP) as aggregator for the sequence to classify.}
\label{tab:action_recognition}
\begin{tabular}{llcc}
\toprule
Model & Modality & Top-1\,$\uparrow$ & Macro-F1\,$\uparrow$ \\
\midrule
ResNet-50~\cite{he2016deep}(MP) & RGB   & 5.0  & 7.0  \\
VideoMAE~\cite{tong2022videomae}         & RGB   & 12.5 & 9.5  \\
\midrule
ResNet-50~\cite{he2016deep}(MP) & Event & \textbf{25.0} & \textbf{20.7} \\
VideoMAE~\cite{tong2022videomae}         & Event & 15.0 & 10.2 \\
\bottomrule
\end{tabular}
\end{subtable}
\end{table}

\minisection{Action Recognition.} Table~\ref{tab:action_recognition} summarizes action recognition on 20 interaction classes. Overall performance is low, reflecting the fine-grained nature of the labels and the strong viewpoint/occlusion variation in the dataset. Interestingly, event-based models outperform their RGB counterparts: a simple ResNet-50 with temporal mean pooling achieves the best Top-1 and Macro-F1, while VideoMAE provides only limited gains on RGB. This suggests that, for these interactions, motion and contact dynamics captured by events are more informative than appearance cues, and that stronger temporal modeling and/or pose-centric representations may be needed to close the gap.

\begin{table}[tb]
\centering
\caption{Zero-shot 2D robot detection AP$_{50}$ (\%) on DHRP. Models are trained on synthetic exocentric RGB data without fine-tuning on DHRP.}
\label{tab:dhrp_zeroshot_det}
\resizebox{\textwidth}{!}{
\begin{tabular}{l rrrrrrrrrrrr}
\toprule
Model & All\,$\uparrow$ & H1\,$\uparrow$ & EVE\,$\uparrow$ & Kepler\,$\uparrow$ & DARwIn\,$\uparrow$ & Atlas\,$\uparrow$ & FIG01\,$\uparrow$ & Toro\,$\uparrow$ & Apollo\,$\uparrow$ & Optimus\,$\uparrow$ & TALOS\,$\uparrow$ & Phoenix\,$\uparrow$ \\
\midrule
Faster R-CNN~\cite{ren2015faster}         & 29.6 & 39.1 & 76.2 & 35.6 & \textbf{71.5} & 56.0 & 20.4 & \textbf{73.6} & 27.1 & 10.5 & 14.2 & 15.5 \\
Def.\,DETR~\cite{zhu2020deformable}       & 24.8 & 38.4 & 40.4 & 17.2 & 53.7 & 54.7 & 28.4 & 54.0 & 30.2 & 13.3 &  4.0 & 10.6 \\
RT-DETR~\cite{zhao2024detrs}              & 44.8 & 99.0 & 77.9 & \textbf{69.0} & 49.8 & 53.5 & 50.5 & 32.8 & \textbf{43.9} & 56.5 & 30.6 & \textbf{16.9} \\
DINO~\cite{zhang2022dino}                 & \textbf{55.4} & 88.3 & \textbf{87.6} & 59.8 & 67.8 & \textbf{80.4} & \textbf{78.3} & 48.2 & 40.1 & \textbf{90.3} & \textbf{66.1} &  3.1 \\
DETR~\cite{carion2020end}                 & 19.1 & 51.6 & 78.6 &  7.7 & 41.6 & 51.7 &  9.1 & 34.1 & 12.3 &  8.8 & 33.5 &  3.1 \\
YOLO11l~\cite{khanam2024yolov11}          & 21.2 & \textbf{99.2} &  0.0 & 51.1 & 17.3 &  1.7 & 49.9 &  0.1 &  7.2 &  0.3 &  3.4 &  8.8 \\
YOLO26l~\cite{sapkota2025yolo26}          & 20.1 & 95.8 & 57.9 & 31.5 & 26.8 & 12.2 & 14.9 &  0.6 & 25.5 &  0.0 &  0.5 &  5.7 \\
\bottomrule
\end{tabular}
}
\end{table}

\minisection{Sim-to-Real.} Table~\ref{tab:dhrp_zeroshot_det} reports zero-shot transfer for 2D robot detection from synthetic exocentric training to real videos from DHRP \cite{heo2024diverse}, without any fine-tuning. While performance drops substantially compared to in-domain testing, highlighting the gap between rendered data and real imagery (sensor noise, backgrounds, and robot appearance), these results show that training on R2RI can already yield non-trivial performance on real data and, crucially, on \emph{previously unseen} robot models with markedly different shapes and scales. Among the evaluated methods, DINO achieves the best mean AP$_{50}$ across robots, with RT-DETR as a strong runner-up. Transfer performance varies widely across platforms: some humanoids (e.g., H1) are detected reliably by most methods, whereas others (e.g., Phoenix) remain challenging, suggesting that appearance and morphology shifts dominate failure modes and motivating future work on domain adaptation and broader synthetic asset coverage.

\section{Conclusions}
In this work, we introduced R2RI, the first large-scale synthetic dataset specifically designed for the Robot-Robot Interaction (RRI) task. The dataset provides a multimodal and multiview representation of interactions between humanoid robots, including synchronized RGB, depth, and event camera data captured from both egocentric and exocentric perspectives. By leveraging a novel human-to-robot mapping algorithm based on URDF models, we generated diverse interaction scenarios derived from human interaction data, enabling the automatic creation of large amounts of annotated training data under varying robots, viewpoints, and illumination conditions.
Beyond dataset creation, we performed a comprehensive experimental analysis across multiple perception tasks, comparing state-of-the-art methods on RGB and event-based modalities. The results highlight the complementary strengths of these sensing modalities: while RGB data remains highly effective under standard conditions, event-based data demonstrates robustness in challenging illumination scenarios.

\minisection{Limitations} R2RI is fully synthetic, so a residual sim-to-real gap remains despite encouraging zero-shot transfer. Moreover, interactions rely on human-to-robot retargeting and currently cover a limited set of humanoid morphologies/classes (mostly pairwise), without modeling higher-level semantics such as intent or task success.

\appendix
\counterwithin{figure}{section}
\counterwithin{table}{section}
\counterwithin{equation}{section}

\section{Reproducibility}
Upon acceptance, we will release the complete R2RI related code together with detailed instructions for training and testing all the considered baseline methods on it.

All models have been trained on a single RTX4090 with 24GB VRAM.

\section{List of Interactions}
\label{sec:appendix_interactions}
The list of 20 interactions used from the INTER-X dataset \cite{xu2024inter} are the following ones, trying to find balance between action with and without collisions: \\
\mylist{Hug, Wave, Hit, Kick, Push, Pull, Slap, Pat on back, \\
Point finger at, Step on foot, Chase, Rock-paper-scissors, \\
Link arms, Bend, Chat, Pat on cheek, Thumb up, \\
Touch head, Imitate, Look back}. 

\begin{figure}[b]
    \centering
    \setlength{\tabcolsep}{2pt}
    \renewcommand{\arraystretch}{1.05}
    \begin{tabular}{cccc}
        \includegraphics[width=0.22\textwidth]{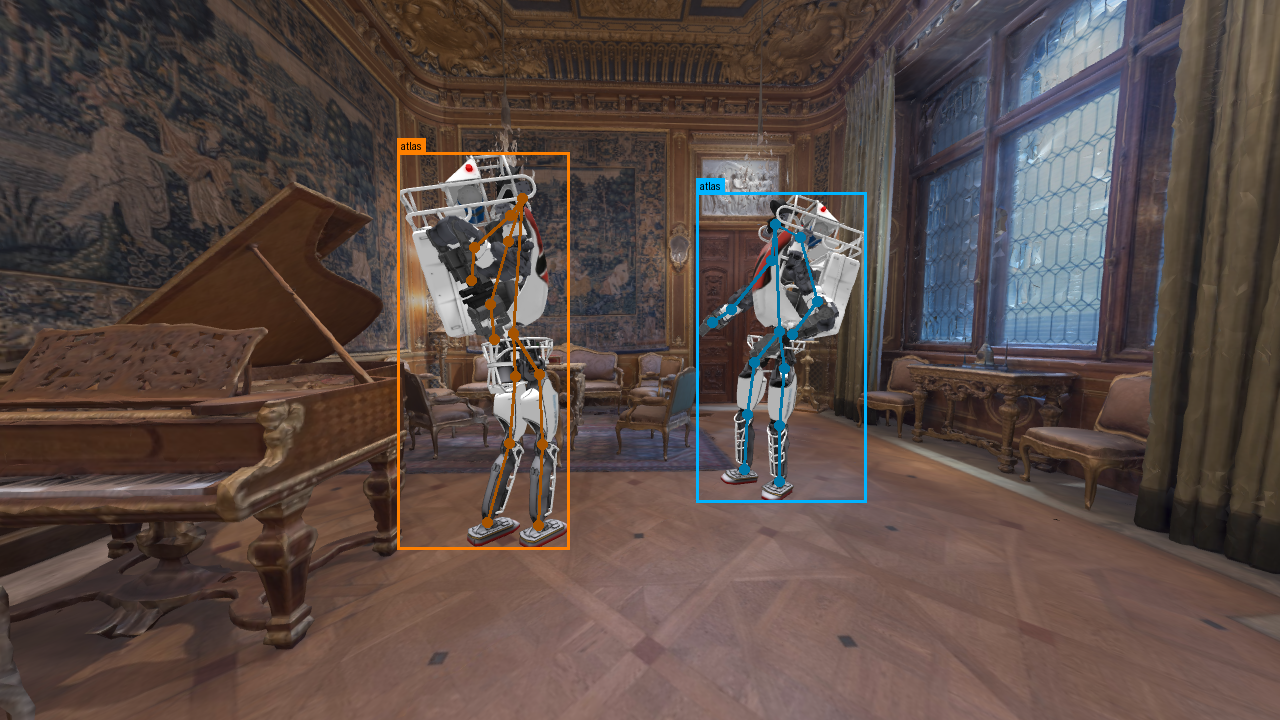} &
        \includegraphics[width=0.22\textwidth]{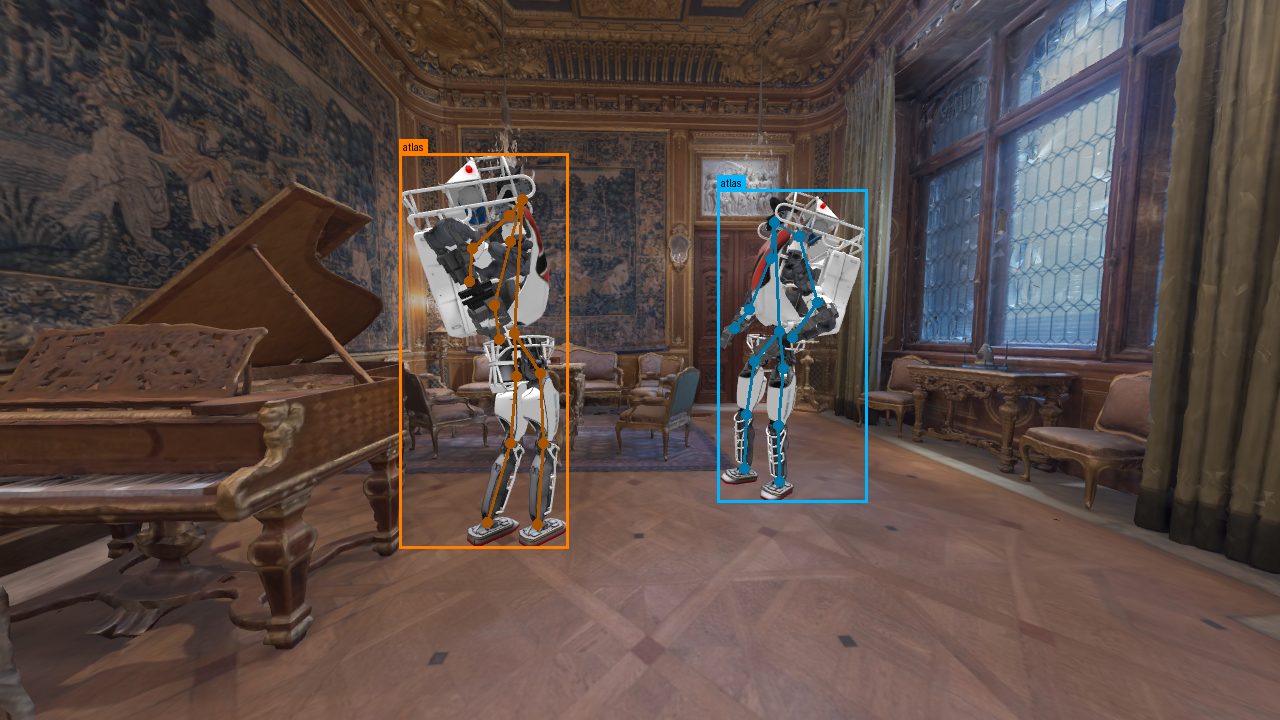} &
        \includegraphics[width=0.22\textwidth]{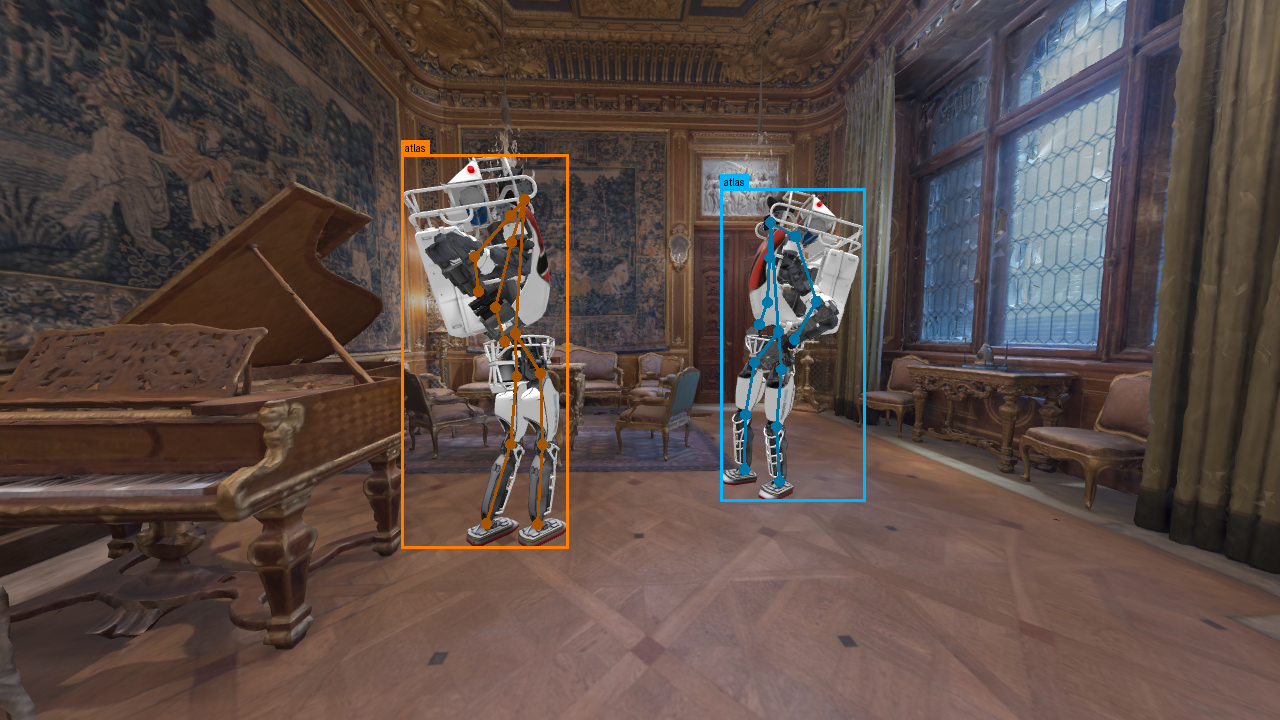} &
        \includegraphics[width=0.22\textwidth]{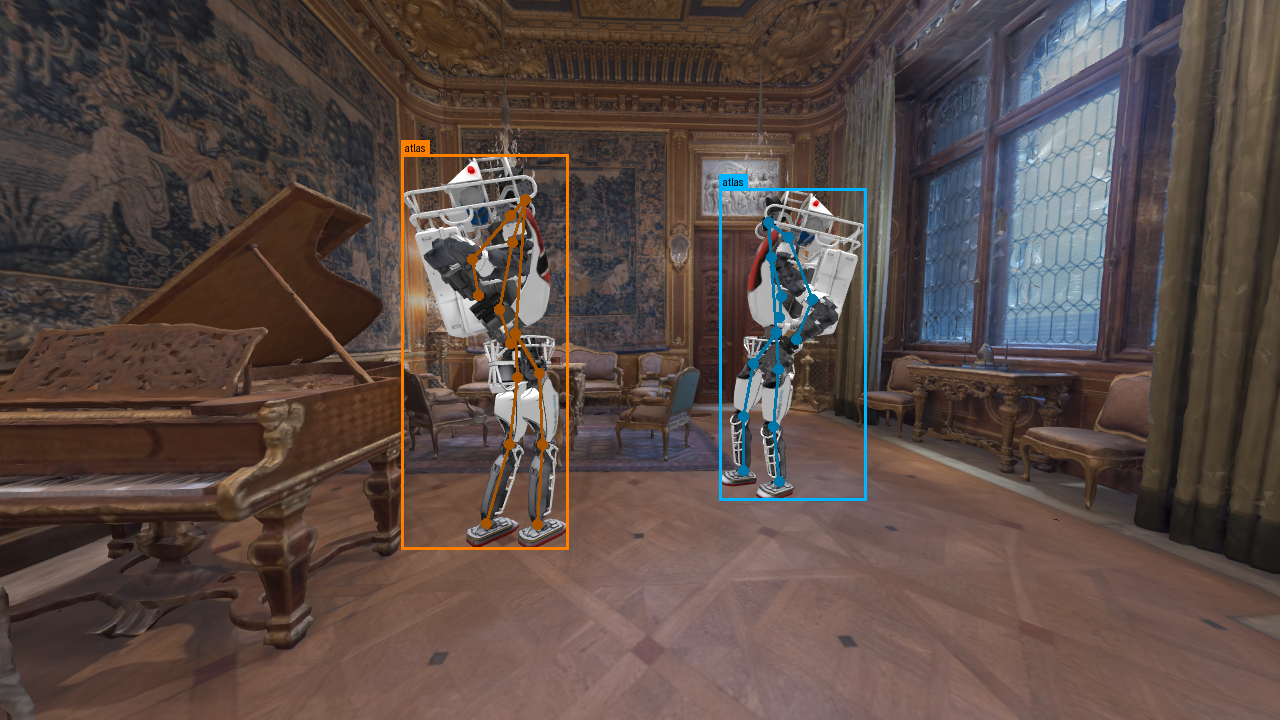} \\
       \includegraphics[width=0.22\textwidth]{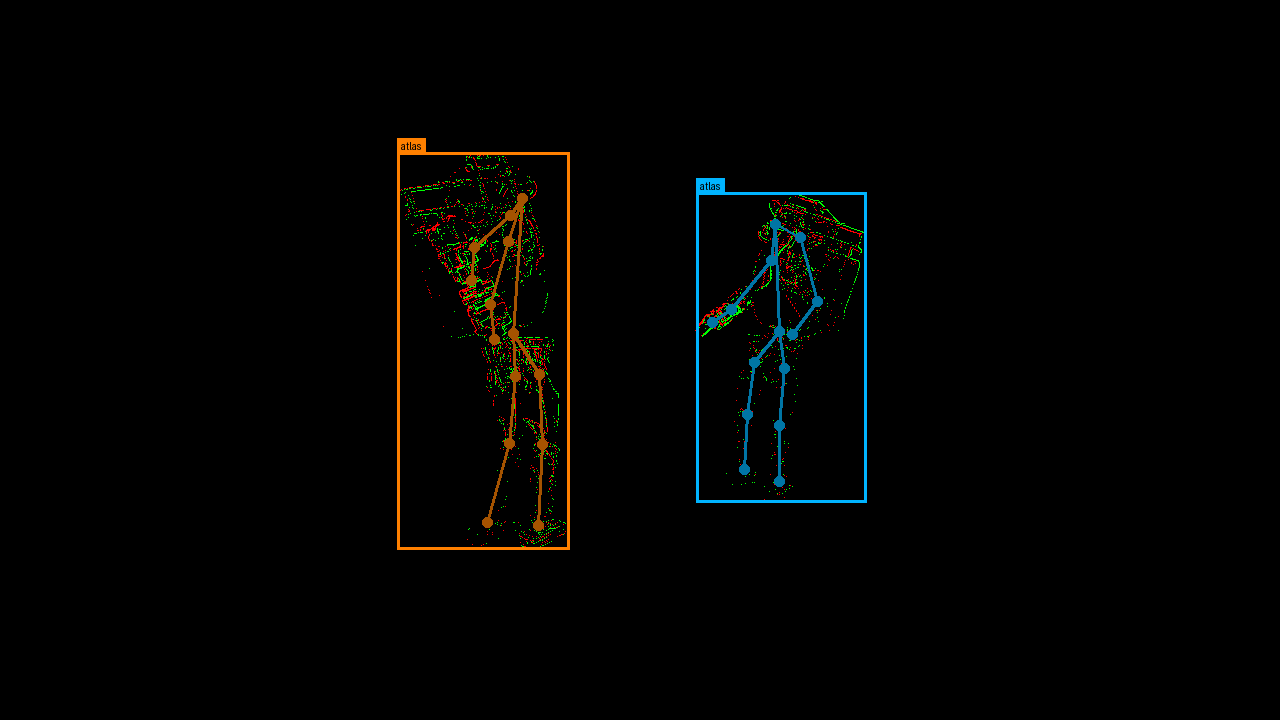} &
        \includegraphics[width=0.22\textwidth]{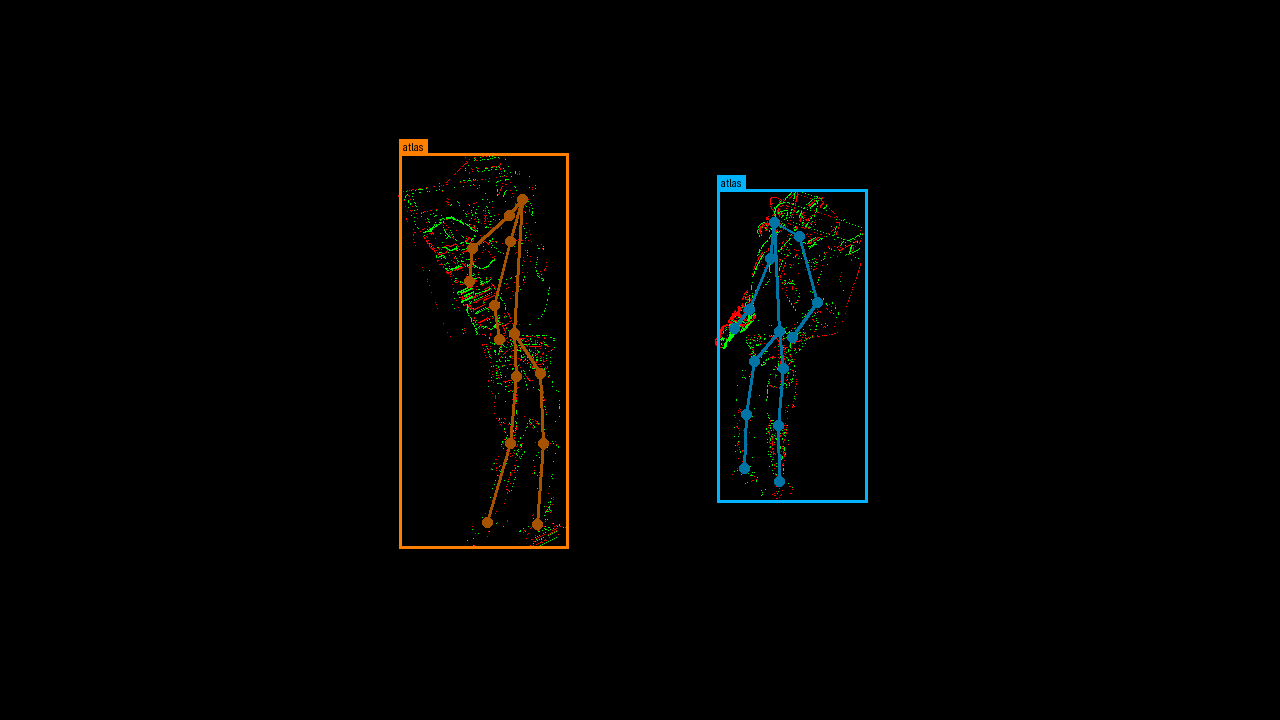} &
        \includegraphics[width=0.22\textwidth]{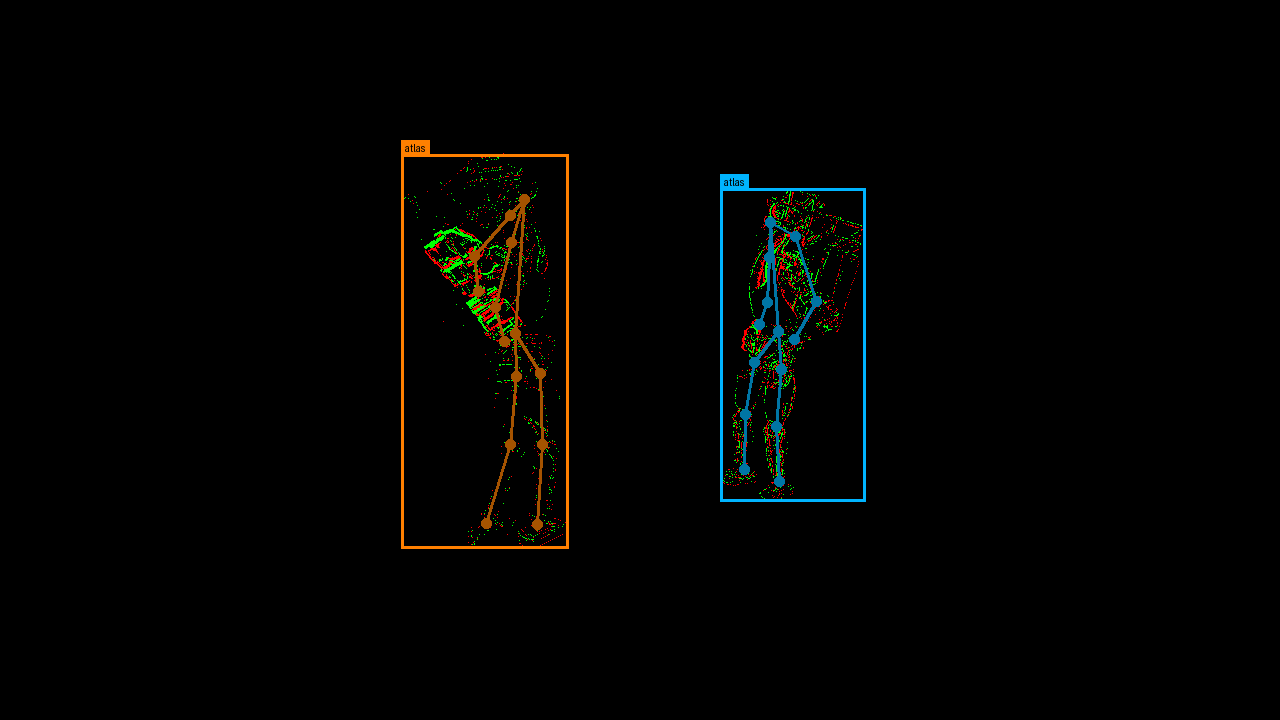} &
        \includegraphics[width=0.22\textwidth]{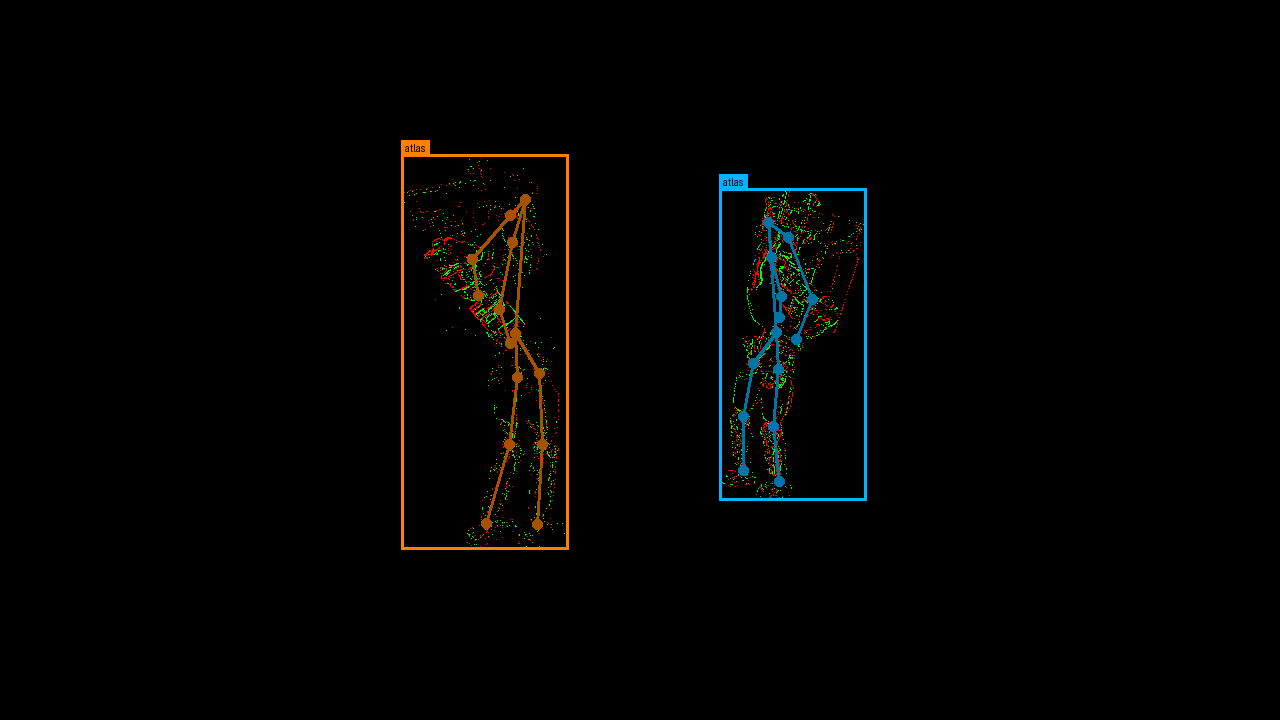} \\
        \includegraphics[width=0.22\textwidth]{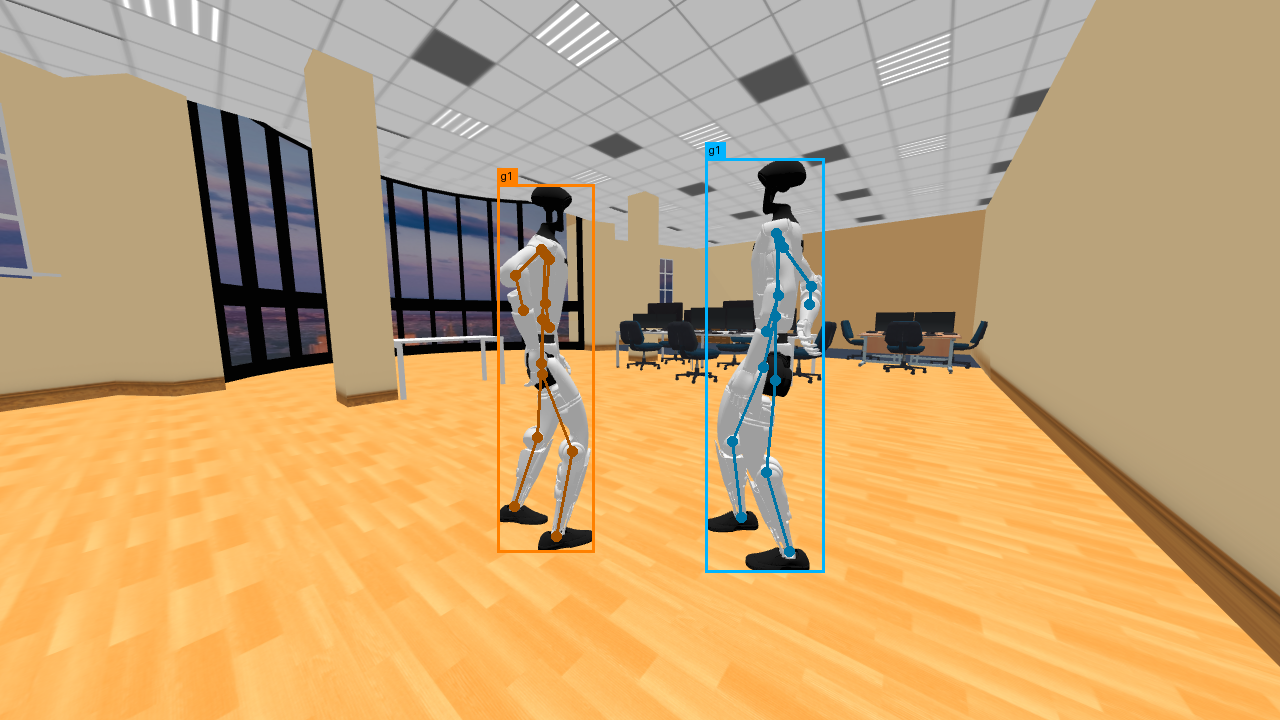} &
        \includegraphics[width=0.22\textwidth]{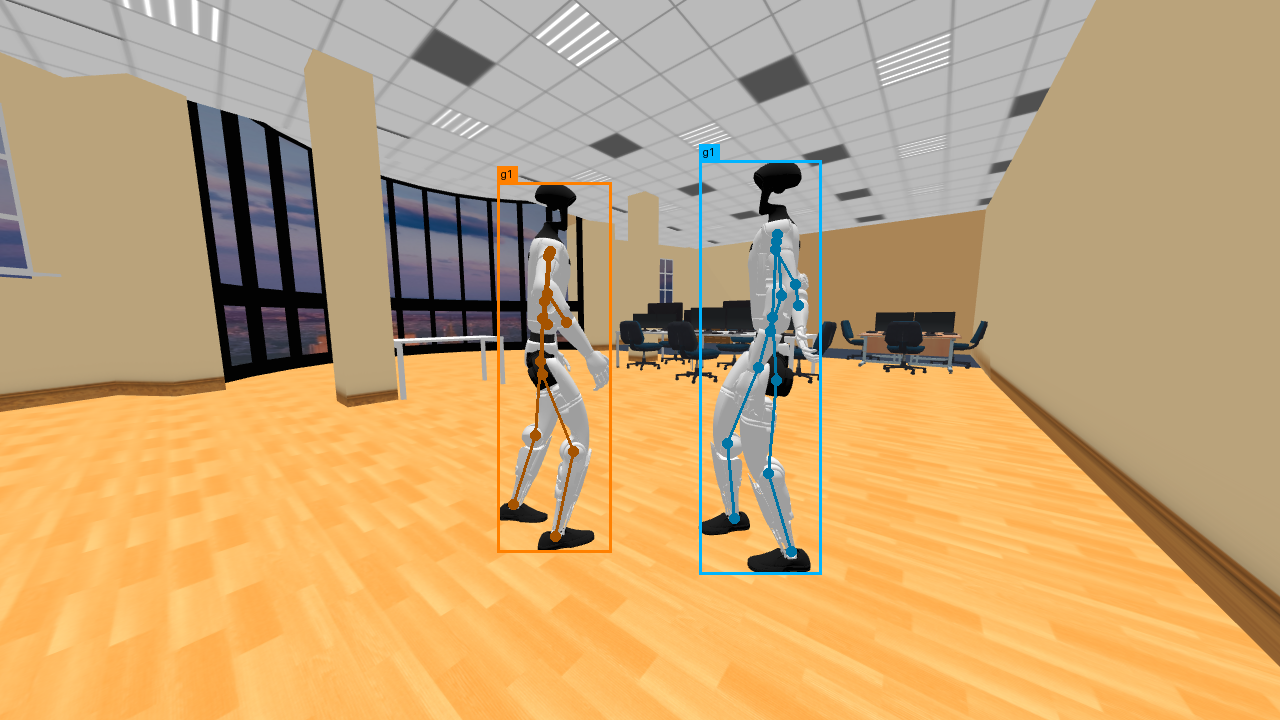} &
        \includegraphics[width=0.22\textwidth]{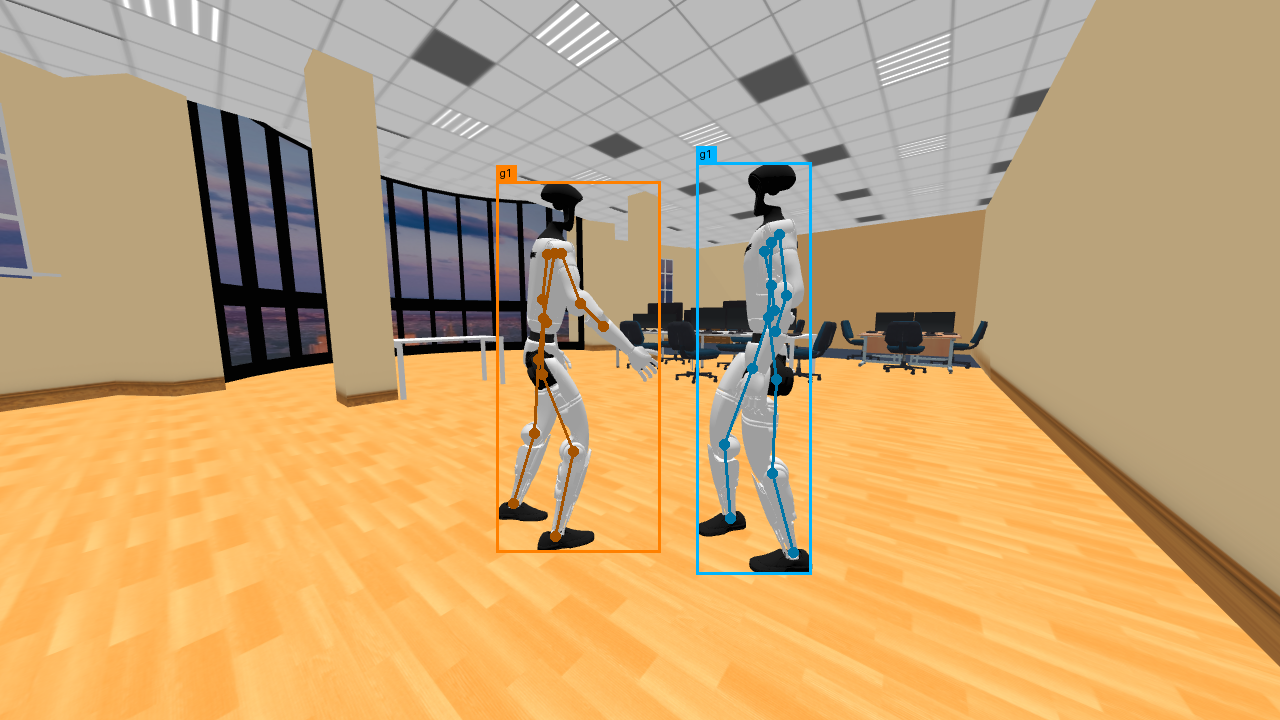} &
        \includegraphics[width=0.22\textwidth]{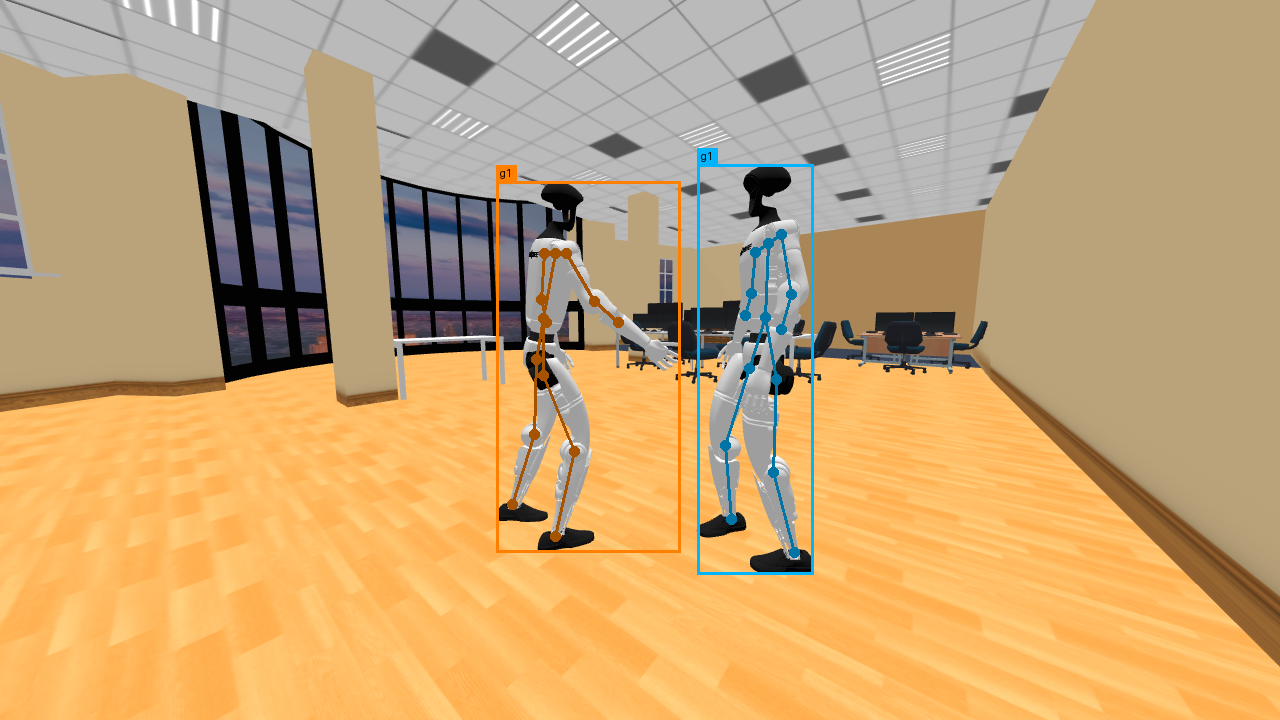} \\
        \includegraphics[width=0.22\textwidth]{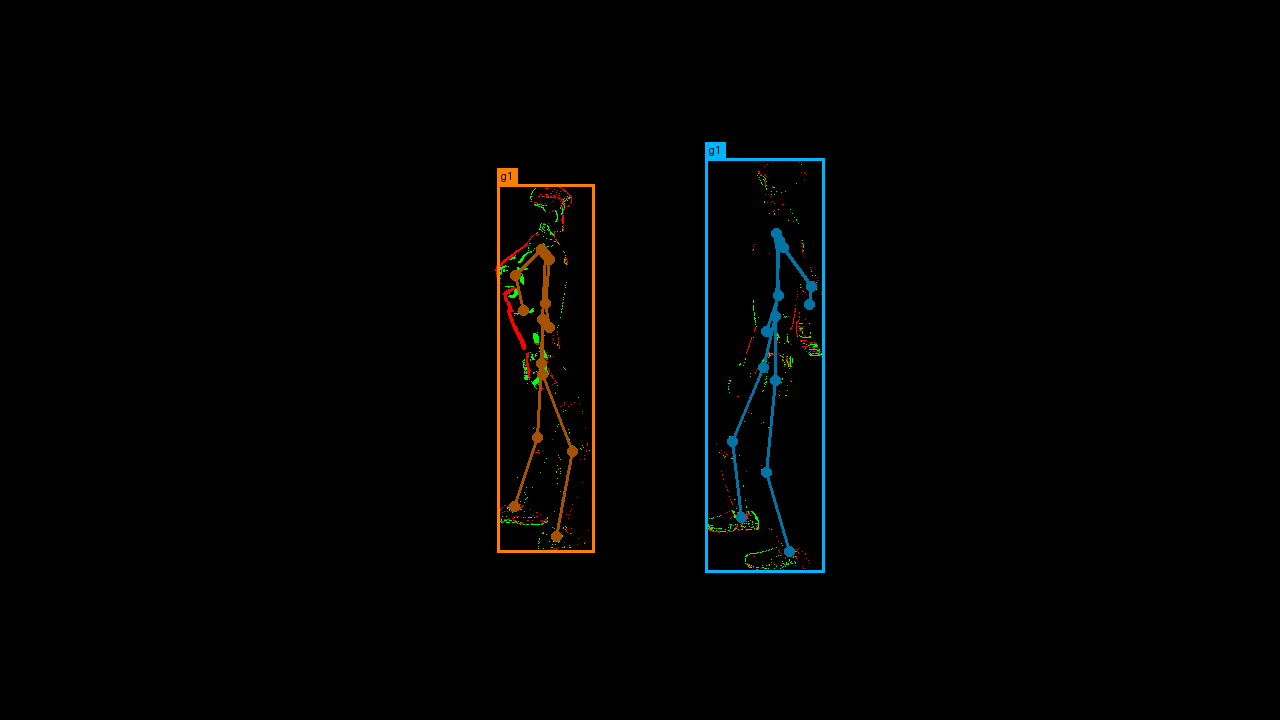} &
        \includegraphics[width=0.22\textwidth]{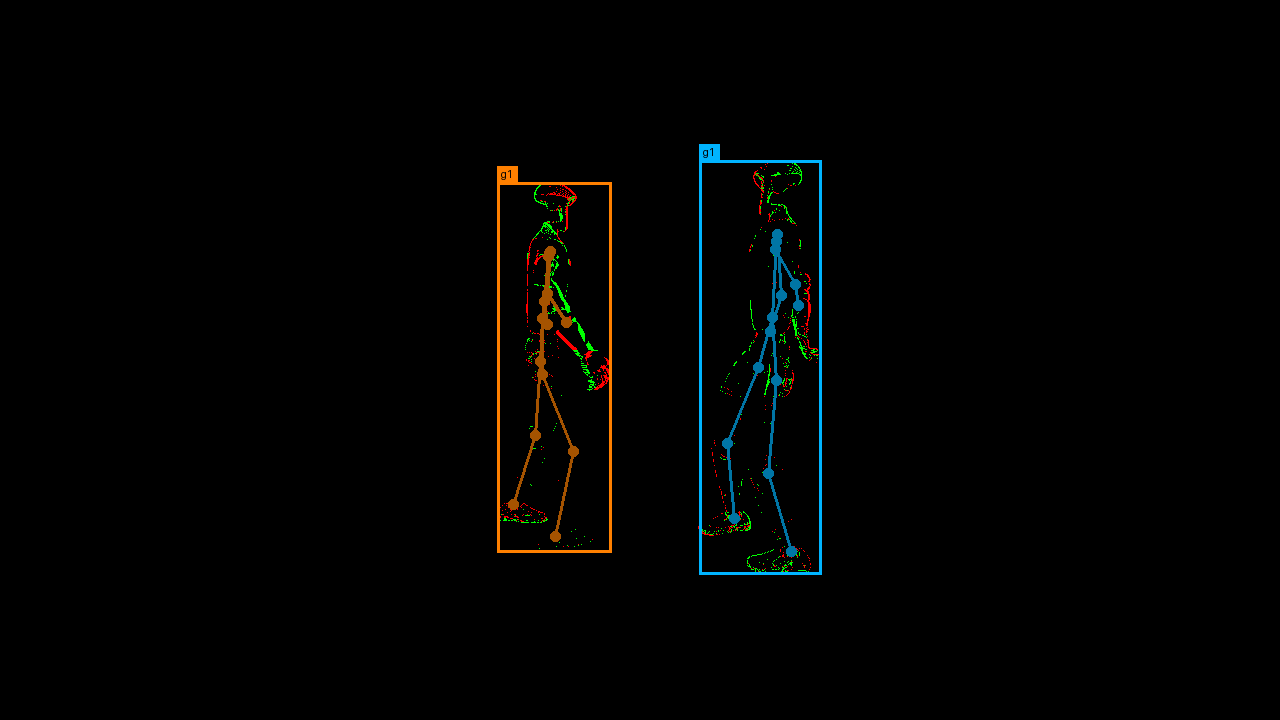} &
        \includegraphics[width=0.22\textwidth]{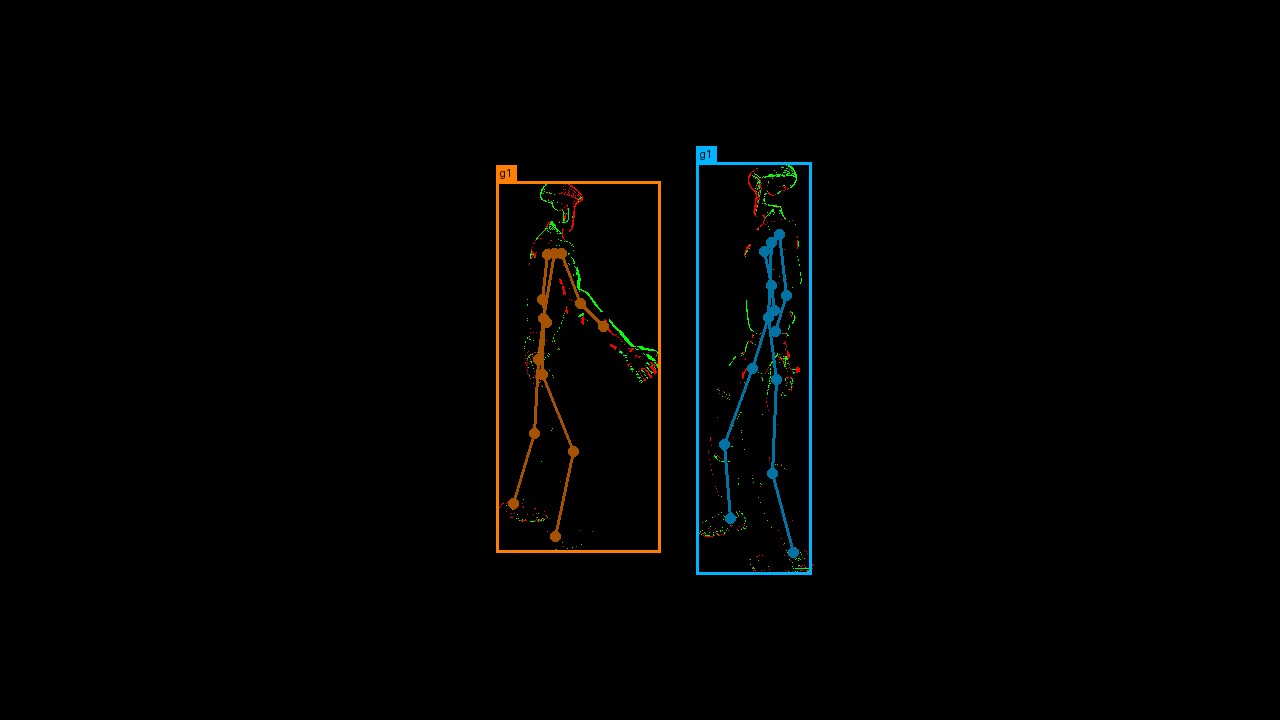} &
        \includegraphics[width=0.22\textwidth]{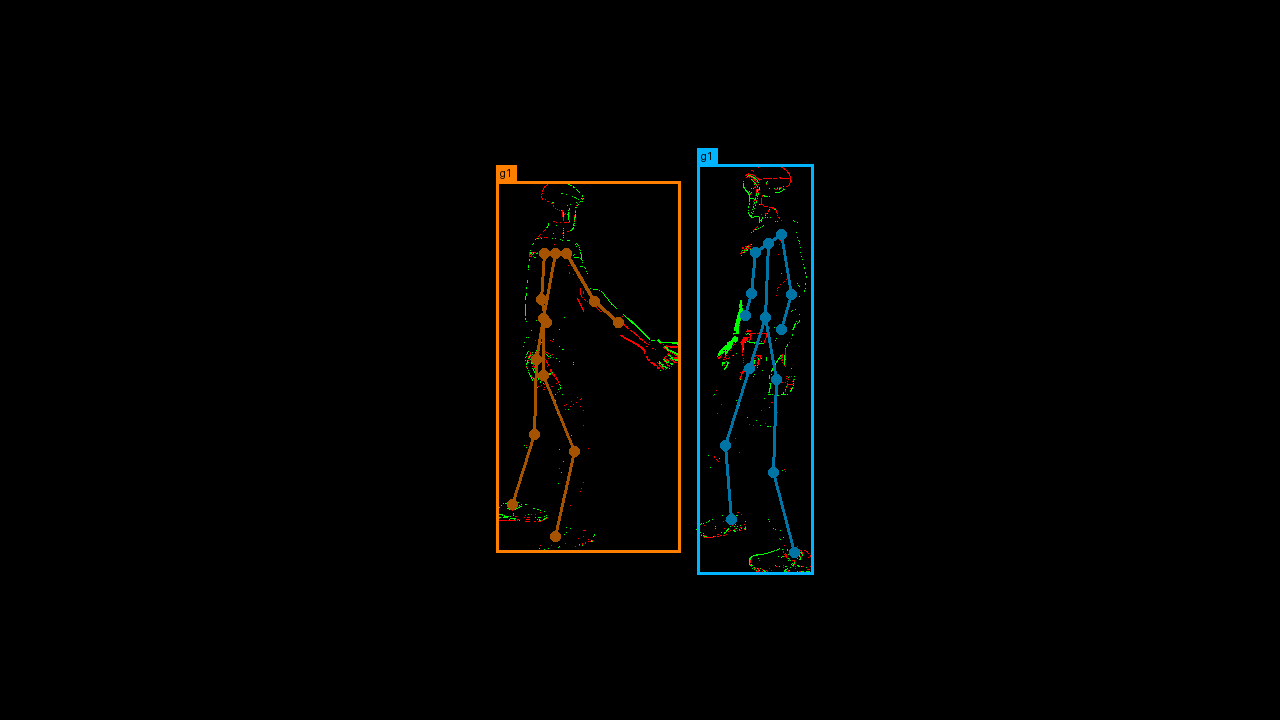} \\
        \textbf{$T_0$} & \textbf{$T_1$} & \textbf{$T_2$} & \textbf{$T_3$} \\
    \end{tabular}
    \caption{Examples of movement sequences, captured each 10 frames. Annotations for both detection and 2D pose are displayed.}
    \label{fig:ego-exo-examples}
\end{figure}

\section{Background 3d Scenes}

We selected 6 different background 3D scenes. Among them four are indoor enviroments and two are outdoor environments.
\subsection{Indoor Environments}
\begin{itemize}
    \item \textbf{Historical Room \cite{drawing_room}:} the Great Drawing Room of The Hallwyl Museum in Stockholm (Sweden)
    \item \textbf{Art Museum \cite{bigazzi2022embodied}:} multiple rooms of the Galleria Estense museum of Modena (Italy)
    \item \textbf{Hospital \cite{queisner2022volumetricor}:} an operating room at Campus Mitte, Charite - Universit atsmedizin, Berlin (Germany) 
    \item \textbf{Office\cite{mersus_office}:} an office of Mersus Technologies (Ireland)
\end{itemize}

\subsection{Outdoor Environments}
\begin{itemize}
    \item \textbf{Artificial City \cite{city_buildings}}: a realistic game a-like urban environment.
    \item \textbf{Borgo \cite{via_gemini}}: one of the main street of a small town in Viterbo (Italy)
\end{itemize}

\section{Robot Models}

Four different robot models were selected based on their wide adoption and URDF availability.
\begin{itemize}
    \item \textbf{Nao robot.} A small humanoid robot designed by Aldebaran Robotics for educational and social research. It is known for limited agility but strong interactive capabilities, tactile sensors, and speech recognition.
    \item \textbf{Unitree G1}. A bipedal humanoid robot designed for research and agile locomotion. It focuses on balance, mobility, and dynamic movement, often used for walking, running, and human-robot interaction experiments.    
    \item \textbf{Atlas v5}. An advanced humanoid robot developed for research and demonstration of dynamic skills. It can walk, jump, and manipulate objects in complex environments.
    \item \textbf{ICub}. A medium-sized humanoid robot designed mainly for cognitive robotics, AI research, and human-robot interaction. It features articulated hands and sophisticated sensors for learning and manipulation.
\end{itemize}
\section{Challenging Lighting Conditions}
\label{sec:appendix_challenge}
We also present the experiments on the complete set of models for the Challenging Lighting Conditions scenario in Tab. \ref{tab:night_split_detection_complete} and Tab. \ref{tab:night_split_pose_complete}. As discussed in the main paper, it can be seen that models based on the event modality consistently retain an overall high mAP, with most models reporting on par-results across different illumination levels. On the contrary, RGB-based models suffer the illumination change to the point that the mAP at minimum brightness drops almost to 0 for all models.
This behavior is consistent for both the detection and the pose estimation tasks.
\begin{figure}
    \centering
    \setlength{\tabcolsep}{1.5pt}
    \renewcommand{\arraystretch}{1.0}
    \small
    \begin{tabular}{@{}cccc@{}}
        \textbf{100\% Lightness} & \textbf{50\% Lightness} & \textbf{25\% Lightness} & \textbf{5\% Lightness} \\[2pt]

        \includegraphics[trim={10cm 4cm 10cm 4cm},clip,width=0.235\textwidth]{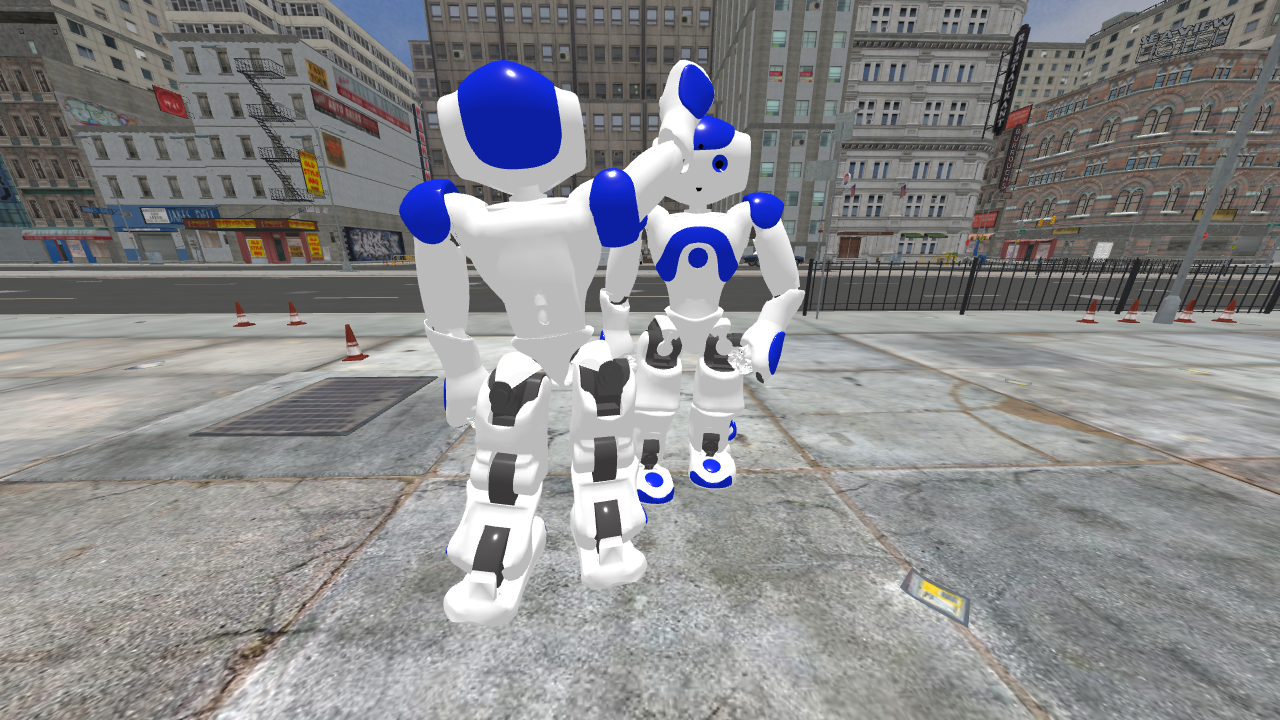} &
        \includegraphics[trim={10cm 4cm 10cm 4cm},clip,width=0.235\textwidth]{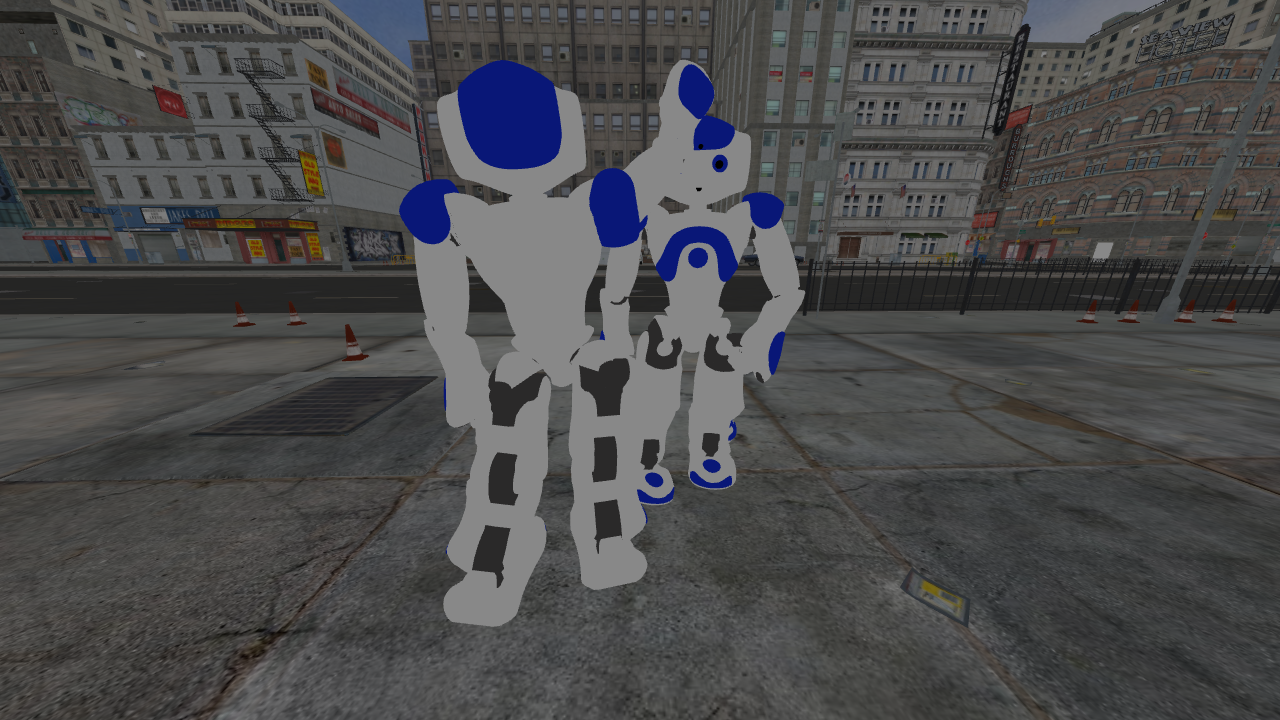} &
        \includegraphics[trim={10cm 4cm 10cm 4cm},clip,width=0.235\textwidth]{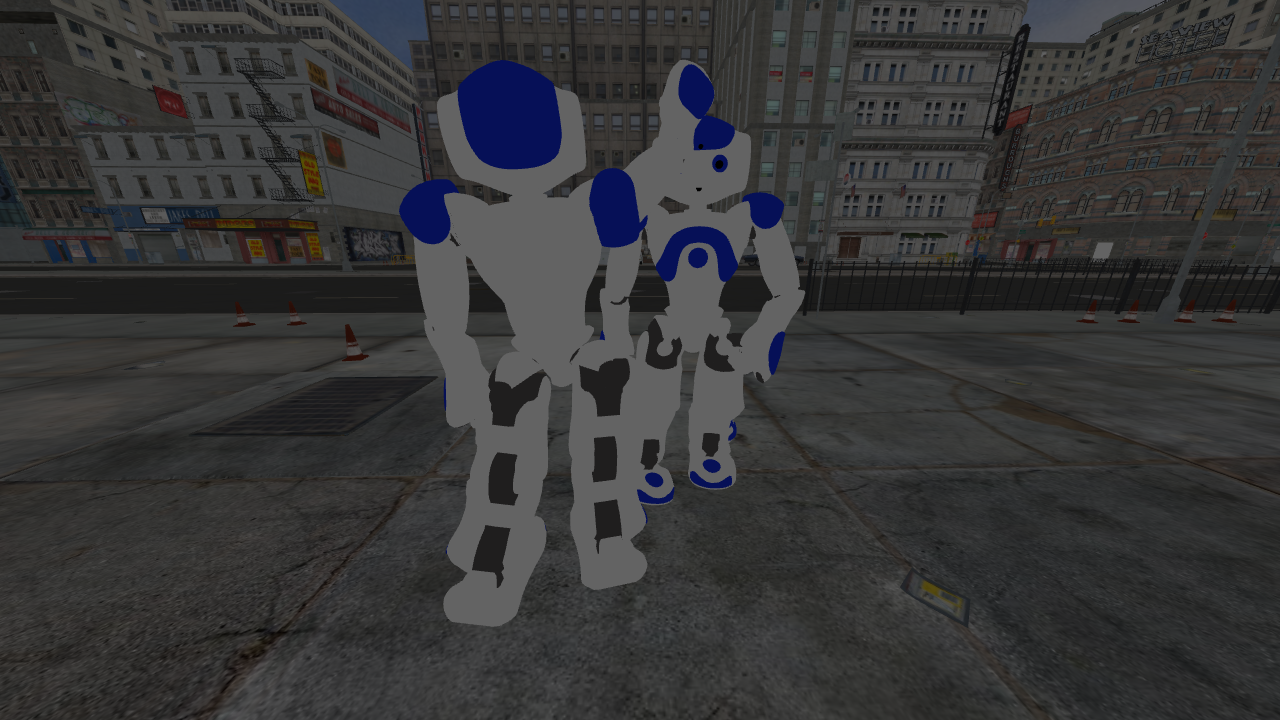} &
        \includegraphics[trim={10cm 4cm 10cm 4cm},clip,width=0.235\textwidth]{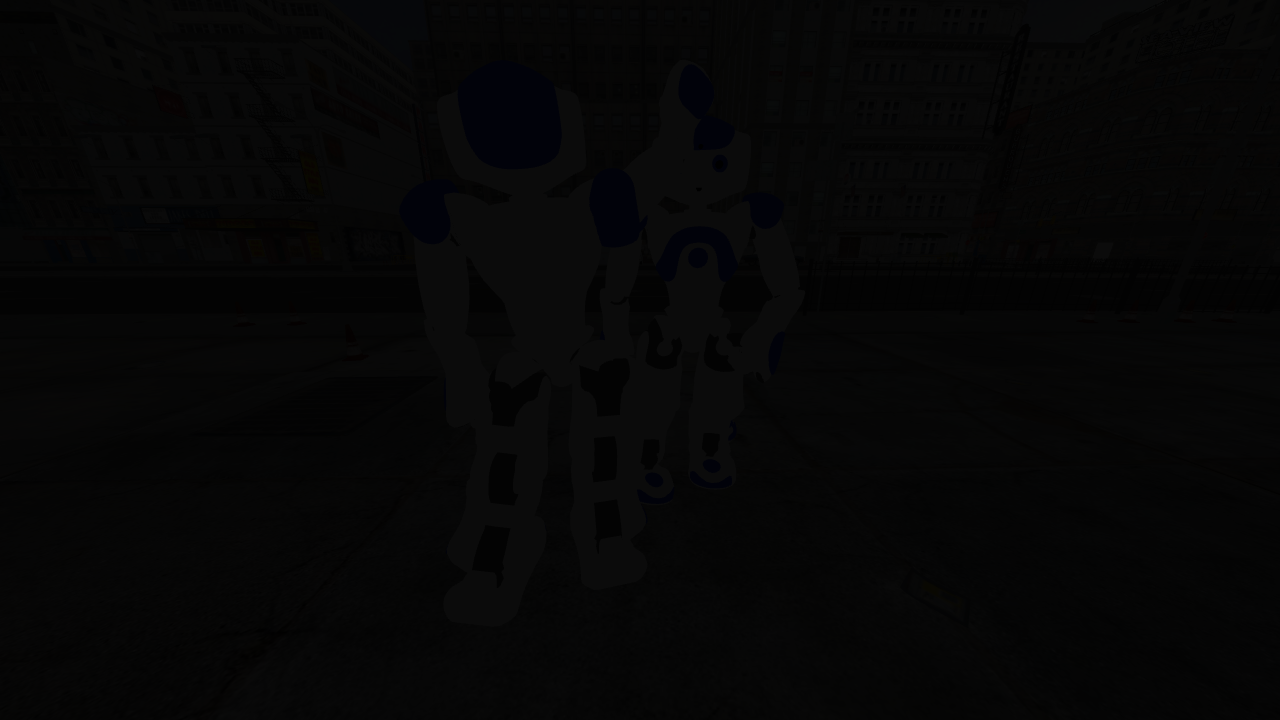} \\[-1pt]

        \multicolumn{4}{c}{\scriptsize\textcolor{gray}{(a) RGB modality}} \\[4pt]

        \includegraphics[trim={10cm 4cm 10cm 4cm},clip,width=0.235\textwidth]{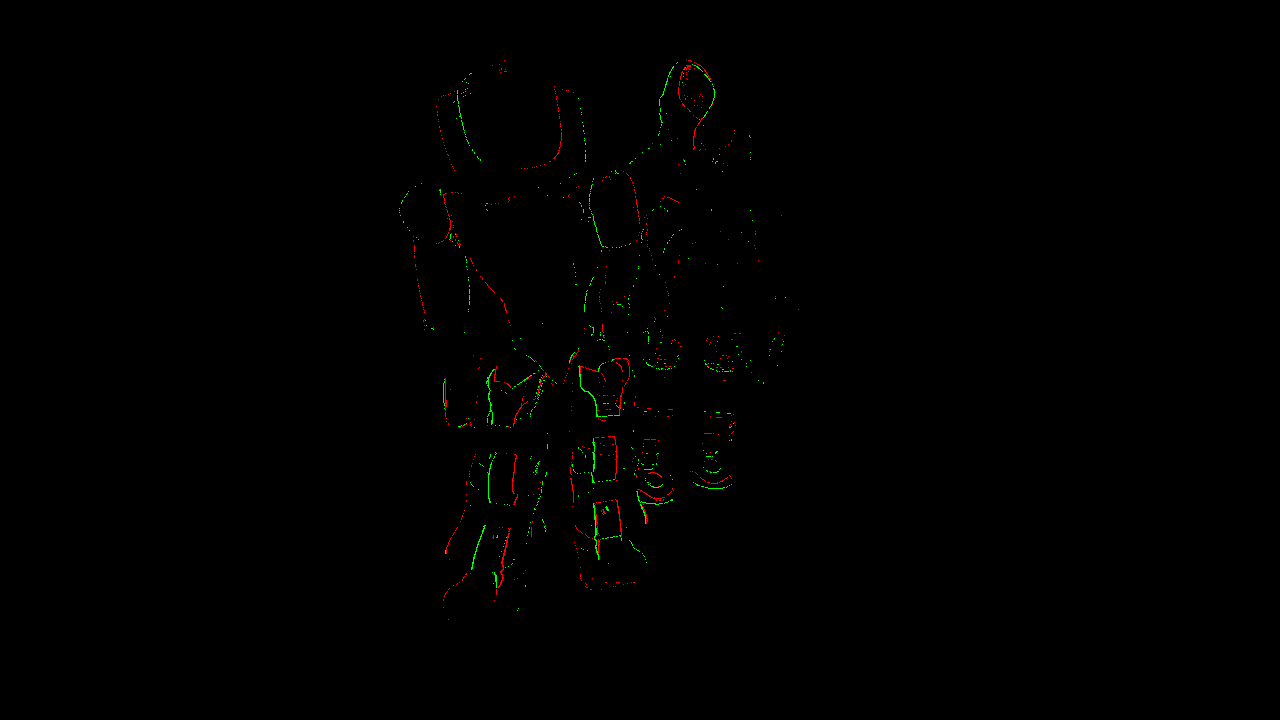} &
        \includegraphics[trim={10cm 4cm 10cm 4cm},clip,width=0.235\textwidth]{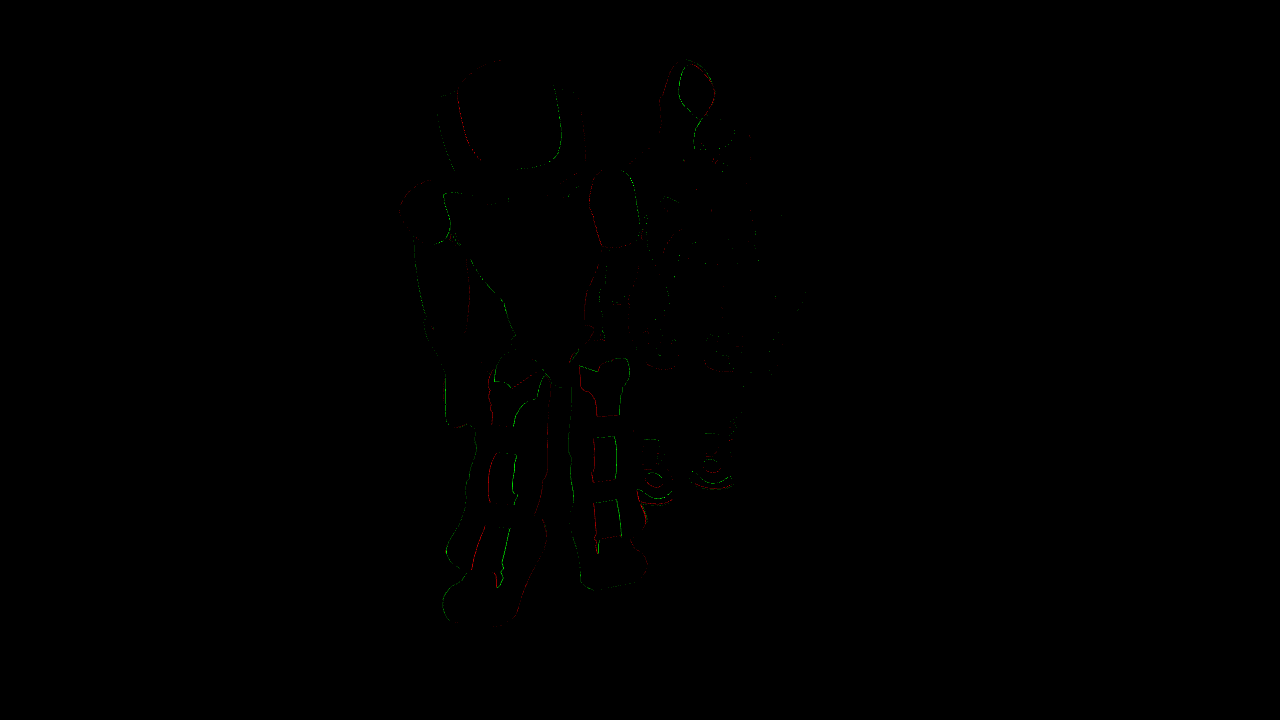} &
        \includegraphics[trim={10cm 4cm 10cm 4cm},clip,width=0.235\textwidth]{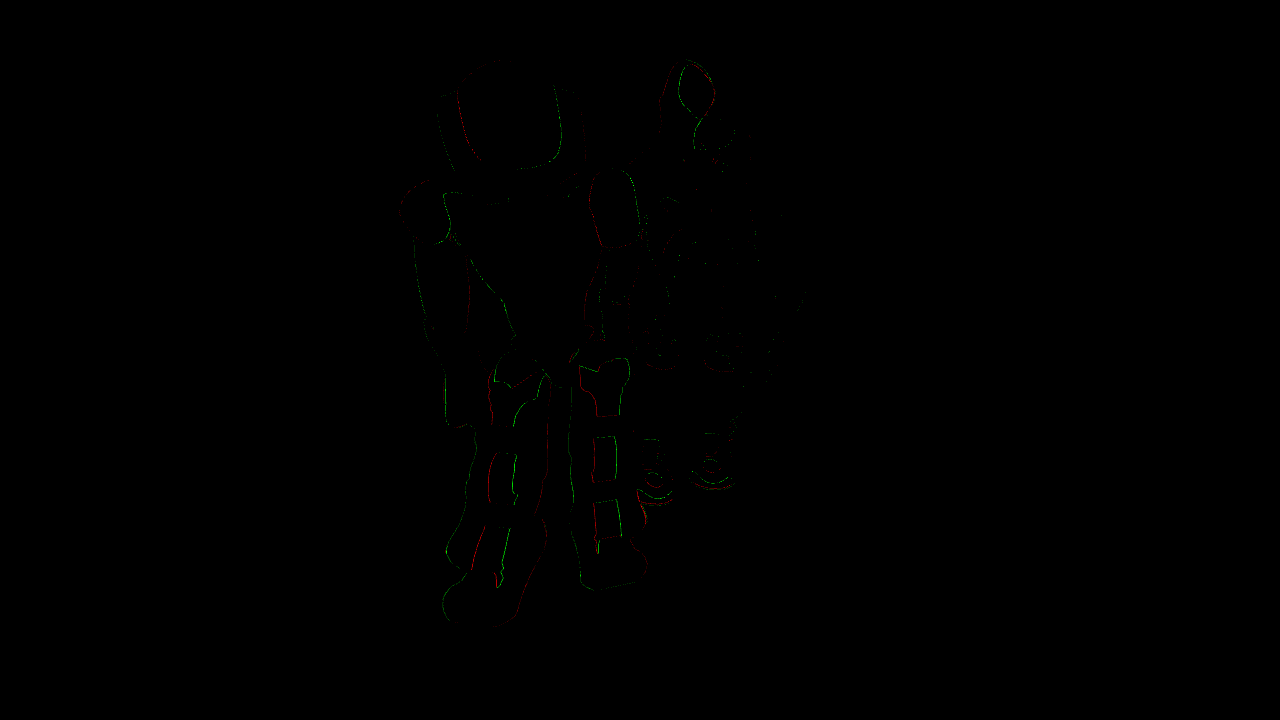} &
        \includegraphics[trim={10cm 4cm 10cm 4cm},clip,width=0.235\textwidth]{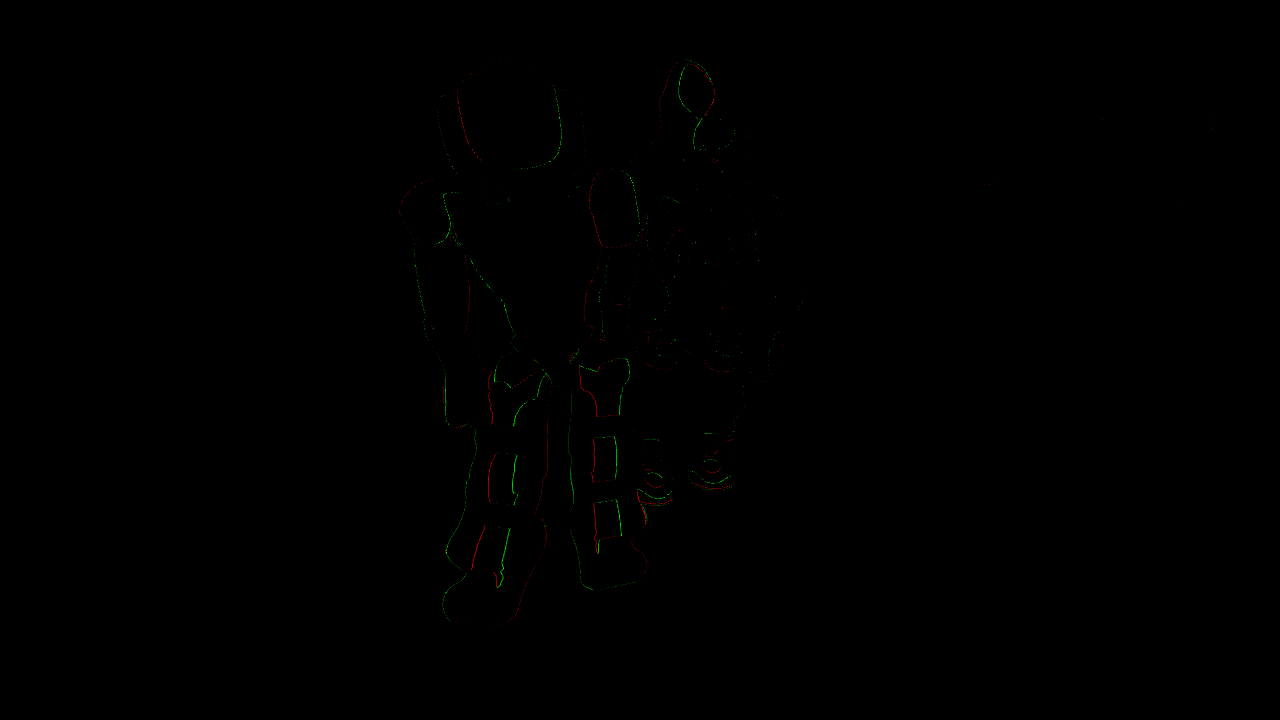} \\[-1pt]

        \multicolumn{4}{c}{\scriptsize\textcolor{gray}{(b) Event modality}} \\
    \end{tabular}

    \vspace{-2pt}
    \caption{\textbf{Robustness to illumination degradation.} Qualitative comparison under progressively reduced visibility (100\%\,\textrightarrow\,5\% lightness). RGB frames become nearly unintelligible at low light levels, while the event modality preserves clear subject silhouettes regardless of illumination.}
    \label{fig:hard_frames}
\end{figure}

\begin{table}
\centering
\footnotesize
\caption{\textbf{Night Split – Detection}. Performance under different illumination levels comparing Event and RGB modalities.}
\label{tab:night_split_detection_complete}
\begin{tabular}{lcccccc}
    \toprule
    & \multicolumn{2}{c}{Half brightness} 
    & \multicolumn{2}{c}{Quarter brightness} 
    & \multicolumn{2}{c}{Min brightness} \\
    \cmidrule(lr){2-3} \cmidrule(lr){4-5} \cmidrule(lr){6-7}
    Model 
    & mAP$_{50}$ & mAP$_{50:95}$ 
    & mAP$_{50}$ & mAP$_{50:95}$ 
    & mAP$_{50}$ & mAP$_{50:95}$ \\
    \midrule
    
    \multicolumn{7}{l}{\textbf{Event Modality}} \\
    \midrule
    Faster R-CNN\cite{ren2015faster}    & 79.6 & 49.7 & 80.2 & 50.2 & 79.7 & 49.0  \\
    Deformable-DETR\cite{zhu2020deformable} & 83.1 & 50.0 & 83.4 & 50.4 & 79.5 & 47.2 \\
    RT-DETR\cite{zhao2024detrs}         & 88.8 & 72.0 & 88.9 & 72.5 & 67.2 & 50.0 \\
    DINO\cite{zhang2022dino}            & 91.3 & 60.4 & 91.8 & 61.0 & 89.6 & 58.6 \\
    DETR\cite{carion2020end}            & 70.1 & 39.3 & 70.5 & 39.7 & 65.0 & 35.6 \\
    YOLO11n\cite{khanam2024yolov11} & 87.1 & 70.7 & 87.6 & 71.6 & 84.3 & 68.1 \\
    YOLO11s\cite{khanam2024yolov11} & 84.5 & 64.1 & 85.0 & 64.7 & 84.2 & 63.7 \\
    YOLO11l\cite{khanam2024yolov11} & 83.4 & 63.1 & 83.8 & 63.8 & 81.2 & 61.3 \\
    YOLO26n\cite{sapkota2025yolo26} & 89.3 & 72.1 & 89.4 & 72.6 & 79.8 & 62.5 \\
    YOLO26s\cite{sapkota2025yolo26} & 92.3 & 75.9 & 92.4 & 76.5 & 87.2 & 71.3 \\
    YOLO26l\cite{sapkota2025yolo26} & 87.7 & 65.3 & 87.8 & 65.8 & 88.1 & 66.0 \\
    \midrule
    
    \multicolumn{7}{l}{\textbf{RGB Modality}} \\
    \midrule
    Faster R-CNN\cite{ren2015faster}    & 94.1  & 69.9 & 91.6  & 62.6  & 4.8  & 1.6  \\
    Deformable-DETR\cite{zhu2020deformable} & 97.9 & 77.7 & 97.6 & 75.4 & 34.3 & 14.1  \\
    RT-DETR\cite{zhao2024detrs}         & 98.5 & 88.2 & 98.1 & 86.4 & 1.4 & 0.4 \\
    DINO\cite{zhang2022dino}            & 98.4 & 80.2 & 98.1 & 78.2 & 38.6 & 18.9 \\
    DETR\cite{carion2020end}            & 97.8 & 68.8 & 95.5 & 65.3 & 1.1 & 0.4\\
    YOLO11n\cite{khanam2024yolov11} & 98.6 & 92.2 & 98.5 & 91.2 & 8.9 & 5.9 \\
    YOLO11s\cite{khanam2024yolov11} & 98.7 & 92.3 & 98.6 & 91.1 & 1.2 & 0.4 \\
    YOLO11l\cite{khanam2024yolov11} & 98.5 & 93.9 & 98.6 & 93.7 & 2.0 & 0.7 \\
    YOLO26n\cite{sapkota2025yolo26} & 98.3 & 90.3 & 98.1 & 89.4 & 0.7 & 0.2 \\
    YOLO26s\cite{sapkota2025yolo26} & 98.2 & 92.8 & 98.3 & 92.7 & 0.7 & 0.2 \\
    YOLO26l\cite{sapkota2025yolo26} & 98.3 & 94.2 & 98.2 & 93.6 & 0.7 & 0.3 \\
    \bottomrule
\end{tabular}
\end{table}

\begin{table}
\centering
\footnotesize
\caption{\textbf{Night Split – Pose Estimation}. Performance under different illumination levels comparing Event and RGB modalities.}
\label{tab:night_split_pose_complete}
\begin{tabular}{lcccccc}
    \toprule
    & \multicolumn{2}{c}{Half brightness} 
    & \multicolumn{2}{c}{Quarter brightness} 
    & \multicolumn{2}{c}{Min brightness} \\
    \cmidrule(lr){2-3} \cmidrule(lr){4-5} \cmidrule(lr){6-7}
    Model 
    & mAP$_{50}$ & mAP$_{50:95}$ 
    & mAP$_{50}$ & mAP$_{50:95}$ 
    & mAP$_{50}$ & mAP$_{50:95}$ \\
    \midrule

    \multicolumn{7}{l}{\textbf{Event Modality}} \\
    \midrule
    RTMPose\cite{jiang2023rtmpose}      & 82.4 & 59.9 & 82.4 & 60.1 & 82.6 & 60.3 \\
    DWPose\cite{yang2023effective}        & 82.3 & 60.1 & 82.3 & 60.3 & 83.4 & 60.9 \\
    HRNET\cite{sun2019deep}        & 61.6 & 39.9 & 62.4 & 40.5 & 62.3 & 42.9 \\
    YOLO11n-pose\cite{khanam2024yolov11} & 76.3 & 45.8 & 76.8 & 46.3 & 78.6 & 48.0 \\
    YOLO11s-pose\cite{khanam2024yolov11} & 79.2 & 50.6 & 79.7 & 51.1 & 81.3 & 52.5 \\
    YOLO11l-pose\cite{khanam2024yolov11} & 79.6 & 52.9 & 79.9 & 53.4 & 79.8 & 53.7 \\
    YOLO26n-pose\cite{sapkota2025yolo26} & 84.3 & 58.6 & 84.7 & 59.0 & 84.1 & 58.4 \\
    YOLO26s-pose\cite{sapkota2025yolo26} & 85.8 & 61.7 & 86.3 & 62.2 & 86.4 & 62.5 \\
    YOLO26l-pose\cite{sapkota2025yolo26} & 88.6 & 67.8 & 89.0 & 68.3 & 86.7 & 66.5 \\
    \midrule

    \multicolumn{7}{l}{\textbf{RGB Modality}} \\
    \midrule
    RTMPose\cite{jiang2023rtmpose}      & 69.5 & 50.8 & 24.1 & 9.9 & 0.0 & 0.0\\
    DWPose\cite{yang2023effective}        & 84.0 & 62.3 & 39.1 & 15.1 & 0.0 & 0.0\\
    HRNET\cite{sun2019deep}        & 91.9 & 67.3 & 87.6 & 57.7 & 1.2 & 0.3 \\
    YOLO11n-pose\cite{khanam2024yolov11} & 95.1 & 82.8 & 93.9 & 80.3 & 7.0 & 2.8 \\
    YOLO11s-pose\cite{khanam2024yolov11} & 95.7 & 85.2 & 94.2 & 82.9 & 2.7 & 0.6 \\
    YOLO11l-pose\cite{khanam2024yolov11} & 97.6 & 89.7 & 97.6 & 89.6 & 10.0 & 3.3 \\
    YOLO26n-pose\cite{sapkota2025yolo26} & 96.2 & 86.2 & 95.2 & 85.0 & 3.1 & 0.5 \\
    YOLO26s-pose\cite{sapkota2025yolo26} & 97.0 & 86.9 & 96.8 & 86.0 & 3.0 & 0.7 \\
    YOLO26l-pose\cite{sapkota2025yolo26} & 97.9 & 90.4 & 98.1 & 90.7 & 1.9 & 0.5 \\
    \bottomrule
\end{tabular}
\end{table}

\section{Qualitatives}
\begin{figure}
    \centering
    \includegraphics[width=\linewidth]{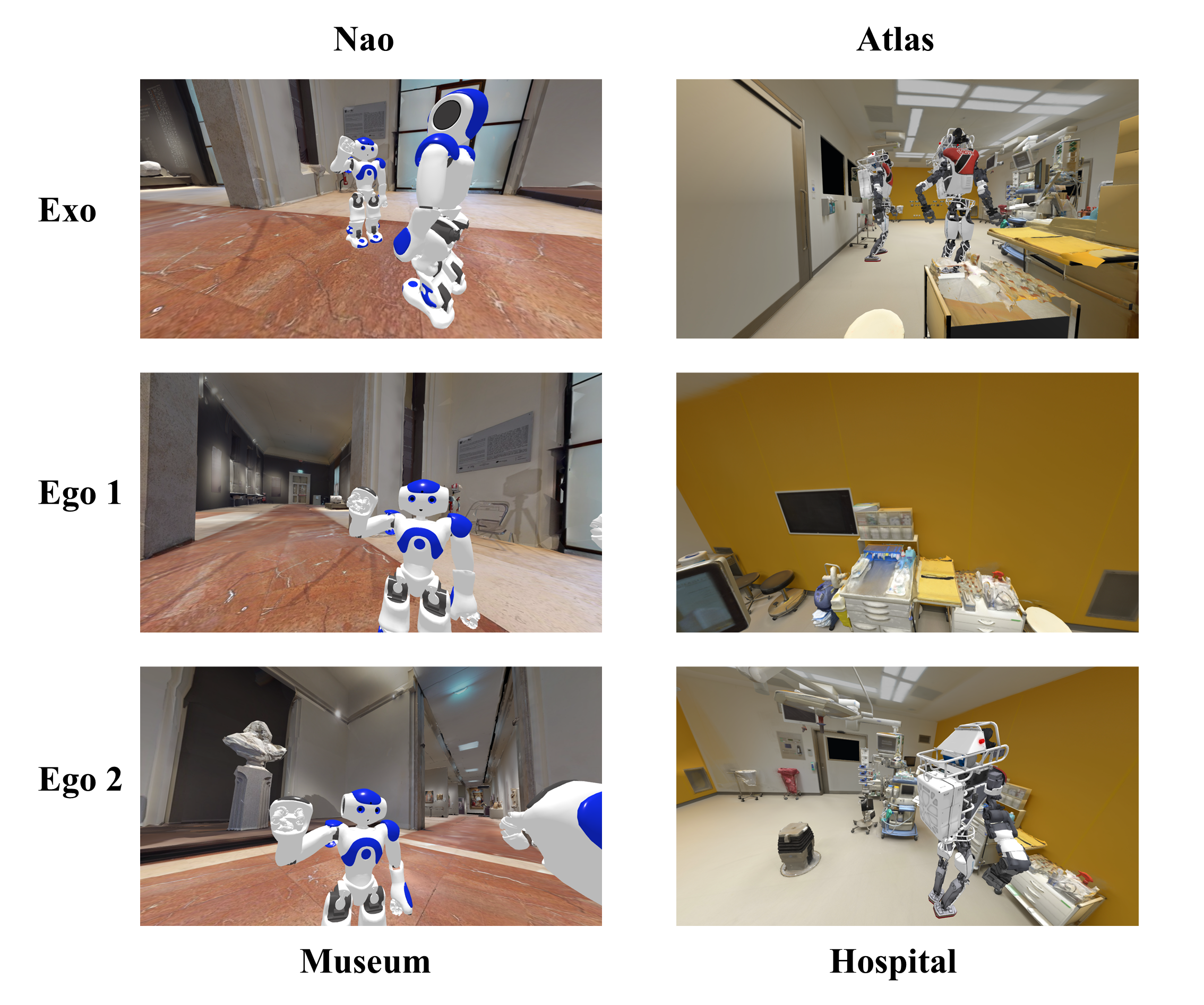}
    \caption{Exocentric and Egocentric (for both robots) views of the same scene in different indoor environments.}
    \label{fig:suppl_qualitative1}
\end{figure}

\begin{figure}
    \centering
    \includegraphics[width=\linewidth]{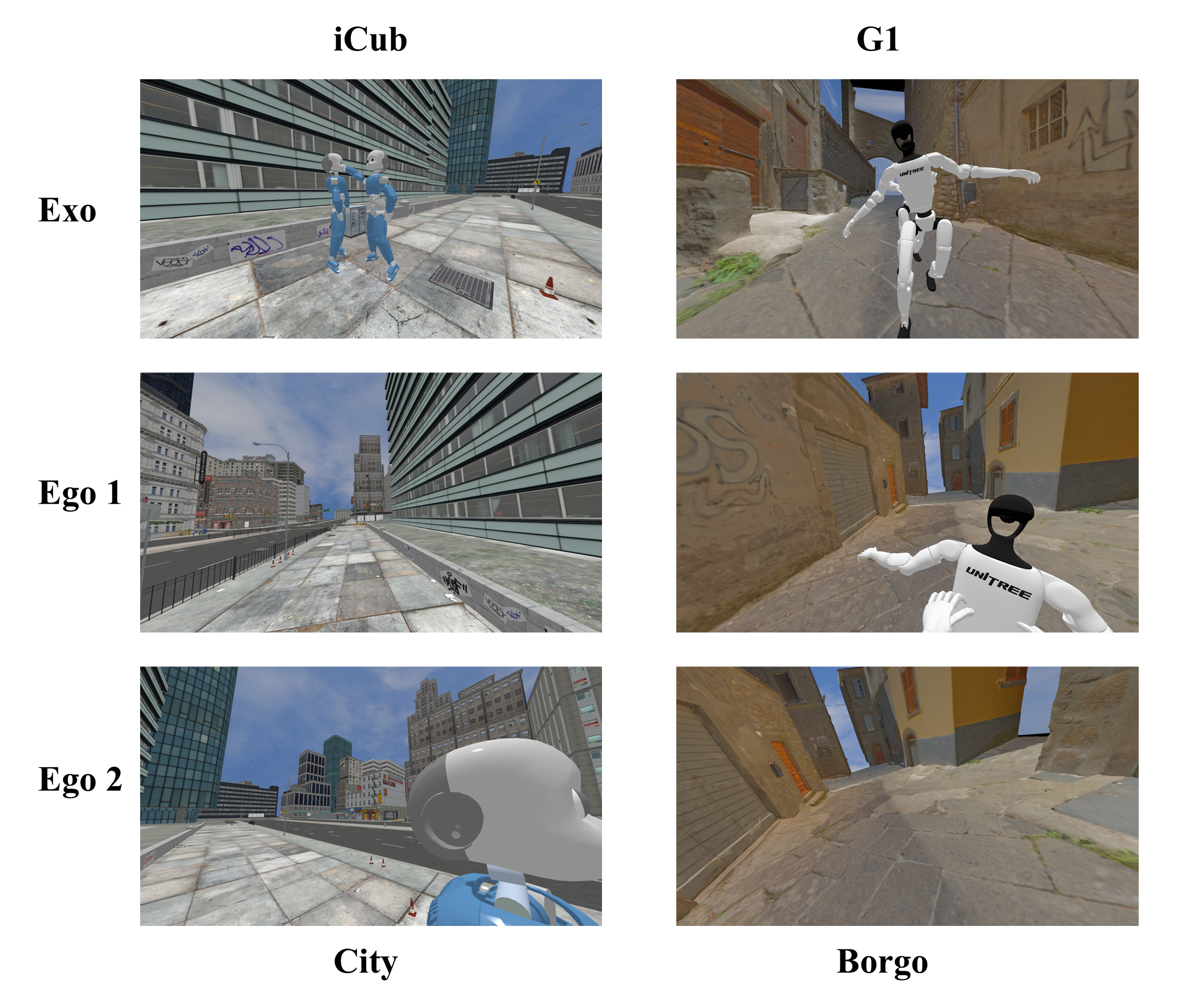}
    \caption{Exocentric and Egocentric (for both robots) views of the same scene in different outdoor environments.}
    \label{fig:suppl_qualitative2}
\end{figure}

In Fig. \ref{fig:suppl_qualitative1} and Fig. \ref{fig:suppl_qualitative2} we report some samples from the dataset, respectively from indoor and outdoor scenes. In the figures, we report both the exocentric view and the two egocentric views (one per robot). It is interesting to notice that egocentric and exocentric view vary considerably and that egocentric views might observe the other robot partially, depending on the development of the action.

\clearpage

\printcredits


\section*{Declaration of generative AI and AI-assisted technologies in the manuscript preparation process}
During the preparation of this work the authors used llms in order to assist with the revision and formatting of the manuscript, which was originally drafted by the authors. After using this tool, the authors reviewed and edited the content as needed and take full responsibility for the content of the published article.

\section*{Data availability}
The R2RI dataset and its annotations for all tasks and modalities are publicly available on Hugging Face at \url{https://huggingface.co/datasets/r2ri/R2RI}. An overview of the dataset with samples is available at \url{https://github.com/MagriniGabriele/R2RI}. The code for the baseline methods will be released upon acceptance.

\section*{Acknowledgements}
This work was partially supported by the project RAINFALL: Recognition Algorithms for Interpreting Neuromorphic-based Facial Actions with Low Latency (piano per lo sviluppo della ricerca PSR 2025, University of Siena). We acknowledge ISCRA for awarding this project access to the LEONARDO supercomputer, owned by the EuroHPC Joint Undertaking, hosted by CINECA (Italy).

\bibliographystyle{elsarticle-num}
\bibliography{bib}

\begin{thebibliography}{10}
\expandafter\ifx\csname url\endcsname\relax
  \def\url#1{\texttt{#1}}\fi
\expandafter\ifx\csname urlprefix\endcsname\relax\def\urlprefix{URL }\fi
\expandafter\ifx\csname href\endcsname\relax
  \def\href#1#2{#2} \def\path#1{#1}\fi

\bibitem{peshkin2001cobot}
M.~A. Peshkin, J.~E. Colgate, W.~Wannasuphoprasit, C.~A. Moore, R.~B.
  Gillespie, P.~Akella, Cobot architecture, IEEE transactions on robotics and
  automation 17~(4) (2001) 377--390.

\bibitem{bicchi2008safety}
A.~Bicchi, M.~A. Peshkin, J.~E. Colgate, Safety for physical human--robot
  interaction, Springer handbook of robotics (2008) 1335--1348.

\bibitem{weiss2011exploring}
A.~Weiss, R.~Buchner, M.~Tscheligi, H.~Fischer, Exploring human-robot
  cooperation possibilities for semiconductor manufacturing, in: 2011
  international conference on collaboration technologies and systems (CTS),
  IEEE, 2011, pp. 173--177.

\bibitem{kolbeinsson2019foundation}
A.~Kolbeinsson, E.~Lagerstedt, J.~Lindblom, Foundation for a classification of
  collaboration levels for human-robot cooperation in manufacturing, Production
  \& Manufacturing Research 7~(1) (2019) 448--471.

\bibitem{simoni20253d}
A.~Simoni, F.~Marchetti, G.~Borghi, F.~Becattini, L.~Seidenari, R.~Vezzani,
  A.~Del~Bimbo, 3d pose nowcasting: Forecast the future to improve the present,
  Computer Vision and Image Understanding 251 (2025) 104233.

\bibitem{dautenhahn2011new}
K.~Dautenhahn, J.~Saunders, New frontiers in human-robot interaction, Vol.~2,
  John Benjamins Publishing, 2011.

\bibitem{geravand2013human}
M.~Geravand, F.~Flacco, A.~De~Luca, Human-robot physical interaction and
  collaboration using an industrial robot with a closed control architecture,
  in: 2013 IEEE international conference on robotics and Automation, IEEE,
  2013, pp. 4000--4007.

\bibitem{kalaitzakis2021fiducial}
M.~Kalaitzakis, B.~Cain, S.~Carroll, A.~Ambrosi, C.~Whitehead, N.~Vitzilaios,
  Fiducial markers for pose estimation: Overview, applications and experimental
  comparison of the artag, apriltag, aruco and stag markers, Journal of
  Intelligent \& Robotic Systems 101~(4) (2021) 71.

\bibitem{fink2012anthropomorphism}
J.~Fink, Anthropomorphism and human likeness in the design of robots and
  human-robot interaction, in: International conference on social robotics,
  Springer, 2012, pp. 199--208.

\bibitem{xu2024inter}
L.~Xu, X.~Lv, Y.~Yan, X.~Jin, S.~Wu, C.~Xu, Y.~Liu, Y.~Zhou, F.~Rao, X.~Sheng,
  et~al., Inter-x: Towards versatile human-human interaction analysis, in:
  Proceedings of the IEEE/CVF Conference on Computer Vision and Pattern
  Recognition (CVPR), 2024, pp. 22260--22271.

\bibitem{quigley2009ros}
M.~Quigley, K.~Conley, B.~Gerkey, J.~Faust, T.~Foote, J.~Leibs, R.~Wheeler,
  A.~Y. Ng, et~al., Ros: an open-source robot operating system, in: ICRA
  workshop on open source software, Vol.~3, Kobe, Japan, 2009, p.~5.

\bibitem{o2024open}
A.~O’Neill, A.~Rehman, A.~Maddukuri, A.~Gupta, A.~Padalkar, A.~Lee,
  A.~Pooley, A.~Gupta, A.~Mandlekar, A.~Jain, et~al., Open x-embodiment:
  Robotic learning datasets and rt-x models: Open x-embodiment collaboration 0,
  in: 2024 IEEE International Conference on Robotics and Automation (ICRA),
  IEEE, 2024, pp. 6892--6903.

\bibitem{zhao2025humanoid}
Z.~Zhao, H.~Jing, X.~Liu, J.~Mao, A.~Jha, H.~Yang, R.~Xue, S.~Zakharor,
  V.~Guizilini, Y.~Wang, Humanoid everyday: A comprehensive robotic dataset for
  open-world humanoid manipulation, 2025, URL https://arxiv.
  org/abs/2510.08807.

\bibitem{li2026robointer}
H.~Li, Z.~Wang, Z.-h. Ding, S.~Yang, Y.~Chen, Y.~Tian, X.~Hu, T.~Wang, D.~Lin,
  F.~Zhao, et~al., Robointer: A holistic intermediate representation suite
  towards robotic manipulation, arXiv preprint arXiv:2602.09973 (2026).

\bibitem{zuo2019craves}
Y.~Zuo, W.~Qiu, L.~Xie, F.~Zhong, Y.~Wang, A.~L. Yuille, Craves: Controlling
  robotic arm with a vision-based economic system, in: Proceedings of the
  IEEE/CVF Conference on Computer Vision and Pattern Recognition (CVPR), 2019,
  pp. 4214--4223.

\bibitem{lee2020camera}
T.~E. Lee, J.~Tremblay, T.~To, J.~Cheng, T.~Mosier, O.~Kroemer, D.~Fox,
  S.~Birchfield, Camera-to-robot pose estimation from a single image, in: 2020
  IEEE International Conference on Robotics and Automation (ICRA), IEEE, 2020,
  pp. 9426--9432.

\bibitem{zampino2022fall}
C.~Zampino, F.~Biancospino, M.~Brienza, F.~Laus, R.~Romano, A.~Pennisi,
  V.~Suriani, D.~D. Bloisi, et~al., Fall detection using nao robot pose
  estimation in robocup spl matches, in: CEUR WORKSHOP PROCEEDINGS, CEUR-WS,
  2022, pp. 88--95.

\bibitem{simoni2022semi}
A.~Simoni, S.~Pini, G.~Borghi, R.~Vezzani, Semi-perspective decoupled heatmaps
  for 3d robot pose estimation from depth maps, IEEE Robotics and Automation
  Letters 7~(4) (2022) 11569--11576.

\bibitem{heo2024diverse}
S.~Heo, Y.~Cho, J.~Park, S.~Cho, Z.~Tsoy, H.~Lim, Y.~Cha, Diverse humanoid
  robot pose estimation from images using only sparse datasets, Applied
  Sciences 14~(19) (2024) 9042.

\bibitem{young2025robots}
J.~Young, J.~Roberts, et~al., Robots watching robots: Human pose estimation
  applied to humanoid robots, in: Proceedings-Australasian Conference on
  Robotics and Automation (ACRA 2025), Australian Robotics and Automation
  Association (ARAA), 2025.

\bibitem{devillers2015multimodal}
L.~Devillers, S.~Rosset, G.~D. Duplessis, M.~A. Sehili, L.~B{\'e}chade,
  A.~Delaborde, C.~Gossart, V.~Letard, F.~Yang, Y.~Yemez, et~al., Multimodal
  data collection of human-robot humorous interactions in the joker project,
  in: 2015 international conference on affective computing and intelligent
  interaction (ACII), IEEE, 2015, pp. 348--354.

\bibitem{celiktutan2017multimodal}
O.~Celiktutan, E.~Skordos, H.~Gunes, Multimodal human-human-robot interactions
  (mhhri) dataset for studying personality and engagement, IEEE Transactions on
  Affective Computing 10~(4) (2017) 484--497.

\bibitem{ben2017ue}
A.~Ben-Youssef, C.~Clavel, S.~Essid, M.~Bilac, M.~Chamoux, A.~Lim, Ue-hri: a
  new dataset for the study of user engagement in spontaneous human-robot
  interactions, in: Proceedings of the 19th ACM international conference on
  multimodal interaction, 2017, pp. 464--472.

\bibitem{canevet2020mummer}
O.~Can{\'e}vet, W.~He, P.~Motlicek, J.-M. Odobez, The mummer data set for robot
  perception in multi-party hri scenarios, in: 2020 29th IEEE International
  Conference on Robot and Human Interactive Communication (RO-MAN), IEEE, 2020,
  pp. 1294--1300.

\bibitem{sampieri2022pose}
A.~Sampieri, G.~M.~D. di~Melendugno, A.~Avogaro, F.~Cunico, F.~Setti,
  G.~Skenderi, M.~Cristani, F.~Galasso, Pose forecasting in industrial
  human-robot collaboration, in: Proceedings of the European Conference on
  Computer Vision (ECCV), Springer, 2022, pp. 51--69.

\bibitem{luo2017robot}
S.~Luo, H.~Lu, J.~Xiao, Q.~Yu, Z.~Zheng, Robot detection and localization based
  on deep learning, in: 2017 Chinese Automation Congress (CAC), IEEE, 2017, pp.
  7091--7095.

\bibitem{garrido2014automatic}
S.~Garrido-Jurado, R.~Mu{\~n}oz-Salinas, F.~J. Madrid-Cuevas, M.~J.
  Mar{\'\i}n-Jim{\'e}nez, Automatic generation and detection of highly reliable
  fiducial markers under occlusion, Pattern Recognition 47~(6) (2014)
  2280--2292.

\bibitem{hasegawa2010development}
H.~Hasegawa, Y.~Mizoguchi, K.~Tadakuma, A.~Ming, M.~Ishikawa, M.~Shimojo,
  Development of intelligent robot hand using proximity, contact and slip
  sensing, in: 2010 IEEE International Conference on Robotics and Automation,
  IEEE, 2010, pp. 777--784.

\bibitem{ren2015faster}
S.~Ren, K.~He, R.~Girshick, J.~Sun, Faster r-cnn: Towards real-time object
  detection with region proposal networks, Advances in neural information
  processing systems 28 (2015).

\bibitem{khanam2024yolov11}
R.~Khanam, M.~Hussain, Yolov11: An overview of the key architectural
  enhancements, arXiv preprint arXiv:2410.17725 (2024).

\bibitem{labbe2021single}
Y.~Labb{\'e}, J.~Carpentier, M.~Aubry, J.~Sivic, Single-view robot pose and
  joint angle estimation via render \& compare, in: Proceedings of the IEEE/CVF
  Conference on Computer Vision and Pattern Recognition (CVPR), 2021, pp.
  1654--1663.

\bibitem{noguchi2022watch}
A.~Noguchi, U.~Iqbal, J.~Tremblay, T.~Harada, O.~Gallo, Watch it move:
  Unsupervised discovery of 3d joints for re-posing of articulated objects, in:
  Proceedings of the IEEE/CVF Conference on Computer Vision and Pattern
  Recognition, 2022, pp. 3677--3687.

\bibitem{simoni2024dspdh}
A.~Simoni, G.~Borghi, L.~Garattoni, G.~Francesca, R.~Vezzani, {D-SPDH}:
  Improving 3d robot pose estimation in sim2real scenario via depth data, IEEE
  Access 12 (2024) 166660--166673.

\bibitem{delbruck2016neuromorophic}
T.~Delbruck, Neuromorophic vision sensing and processing, in: 2016 46Th
  european solid-state device research conference (ESSDERC), IEEE, 2016, pp.
  7--14.

\bibitem{posch2014retinomorphic}
C.~Posch, T.~Serrano-Gotarredona, B.~Linares-Barranco, T.~Delbruck,
  Retinomorphic event-based vision sensors: bioinspired cameras with spiking
  output, Proceedings of the IEEE 102~(10) (2014) 1470--1484.

\bibitem{lichtsteiner2008128}
P.~Lichtsteiner, C.~Posch, T.~Delbruck, A $128\times128$ 120 {dB} 15 $\mu$s
  latency asynchronous temporal contrast vision sensor, IEEE journal of
  solid-state circuits 43~(2) (2008) 566--576.

\bibitem{event_ijprai}
S.~Pini, G.~Borghi, R.~Vezzani, Video frame synthesis combining conventional
  and event cameras, International Journal of Pattern Recognition and
  Artificial Intelligence 35~(16) (2021) 2160013.

\bibitem{gallego2020event}
G.~Gallego, T.~Delbr{\"u}ck, G.~Orchard, C.~Bartolozzi, B.~Taba, A.~Censi,
  S.~Leutenegger, A.~J. Davison, J.~Conradt, K.~Daniilidis, et~al., Event-based
  vision: A survey, IEEE transactions on pattern analysis and machine
  intelligence 44~(1) (2020) 154--180.

\bibitem{magrini2025drone}
G.~Magrini, L.~Berlincioni, F.~Becattini, L.~Cultrera, P.~Pala, Drone detection
  with event cameras, in: 2025 IEEE/CVF International Conference on Computer
  Vision Workshops (ICCVW), IEEE, 2025, pp. 4762--4773.

\bibitem{seok2020robust}
H.~Seok, J.~Lim, Robust feature tracking in dvs event stream using b{\'e}zier
  mapping, in: Proceedings of the IEEE/CVF Winter Conference on Applications of
  Computer Vision, 2020, pp. 1658--1667.

\bibitem{renner2019event}
A.~Renner, M.~Evanusa, Y.~Sandamirskaya, Event-based attention and tracking on
  neuromorphic hardware, in: 2019 IEEE/CVF Conference on Computer Vision and
  Pattern Recognition Workshops (CVPRW), IEEE, 2019, pp. 1709--1716.

\bibitem{magrini2025fred}
G.~Magrini, N.~Marini, F.~Becattini, L.~Berlincioni, N.~Biondi, P.~Pala, A.~D.
  Bimbo, Fred: The florence rgb-event drone dataset, in: Proceedings of the
  33rd ACM International conference on multimedia, 2025, pp. 13170--13176.

\bibitem{liu2023motion}
B.~Liu, C.~Xu, W.~Yang, H.~Yu, L.~Yu, Motion robust high-speed light-weighted
  object detection with event camera, IEEE Transactions on Instrumentation and
  Measurement 72 (2023) 1--13.

\bibitem{mitrokhin2018event}
A.~Mitrokhin, C.~Ferm{\"u}ller, C.~Parameshwara, Y.~Aloimonos, Event-based
  moving object detection and tracking, in: 2018 IEEE/RSJ International
  Conference on Intelligent Robots and Systems (IROS), IEEE, 2018, pp. 1--9.

\bibitem{magrini2026pepr}
G.~Magrini, F.~Becattini, N.~Biondi, P.~Pala, Pepr: Privileged event-based
  predictive regularization for domain generalization, in: Proceedings of the
  IEEE/CVF Conference on Computer Vision and Pattern Recognition (CVPR)
  Findings, 2026, pp. 3209--3219.

\bibitem{magrini2024neuromorphic}
G.~Magrini, F.~Becattini, P.~Pala, A.~Del~Bimbo, A.~Porta, Neuromorphic drone
  detection: an event-rgb multimodal approach, in: European Conference on
  Computer Vision, Springer, 2024, pp. 259--275.

\bibitem{shariff2023neuromorphic}
W.~Shariff, M.~S. Dilmaghani, P.~Kielty, J.~Lemley, M.~A. Farooq, F.~Khan,
  P.~Corcoran, Neuromorphic driver monitoring systems: A computationally
  efficient proof-of-concept for driver distraction detection, IEEE Open
  Journal of Vehicular Technology 4 (2023) 836--848.

\bibitem{becattini2025neuromorphic}
F.~Becattini, L.~Berlincioni, L.~Cultrera, A.~Del~Bimbo, Neuromorphic face
  analysis: A survey, Pattern Recognition Letters 187 (2025) 42--48.

\bibitem{mueggler2018event}
E.~Mueggler, Event-based vision for high-speed robotics, Gesellschaft f{\"u}r
  Informatik eV, 2018.

\bibitem{mahlknecht2022exploring}
F.~Mahlknecht, D.~Gehrig, J.~Nash, F.~M. Rockenbauer, B.~Morrell, J.~Delaune,
  D.~Scaramuzza, Exploring event camera-based odometry for planetary robots,
  IEEE Robotics and Automation Letters 7~(4) (2022) 8651--8658.

\bibitem{pyrender}
M.~Matl, Pyrender: library for physically-based rendering and visualization,
  \url{https://github.com/mmatl/pyrender}, version 0.1.45 (2018).

\bibitem{smplx}
G.~Pavlakos, V.~Choutas, N.~Ghorbani, T.~Bolkart, A.~A.~A. Osman, D.~Tzionas,
  M.~J. Black, Expressive body capture: {3D} hands, face, and body from a
  single image, in: Proceedings of the IEEE/CVF Conference on Computer Vision
  and Pattern Recognition (CVPR), 2019, pp. 10975--10985.

\bibitem{Joshy2024}
A.~J. Joshy, J.~T. Hwang, Pyslsqp: A transparent python package for the slsqp
  optimization algorithm modernized with utilities for visualization and
  post-processing, Journal of Open Source Software 9~(103) (2024) 7246.
\newblock \href {https://doi.org/10.21105/joss.07246}
  {\path{doi:10.21105/joss.07246}}.

\bibitem{Rebecq18corl}
H.~Rebecq, D.~Gehrig, D.~Scaramuzza, {ESIM}: an open event camera simulator,
  Conf. on Robotics Learning (CoRL) (Oct. 2018).

\bibitem{reda2022film}
F.~Reda, J.~Kontkanen, E.~Tabellion, D.~Sun, C.~Pantofaru, B.~Curless, Film:
  Frame interpolation for large motion, in: European Conference on Computer
  Vision, Springer, 2022, pp. 250--266.

\bibitem{hu2021v2e}
Y.~Hu, S.-C. Liu, T.~Delbruck, v2e: From video frames to realistic dvs events,
  in: 2021 IEEE/CVF Conference on Computer Vision and Pattern Recognition
  Workshops (CVPRW), IEEE, 2021, pp. 1312--1321.

\bibitem{LinMBHPRDZ14}
T.-Y. Lin, M.~Maire, S.~Belongie, J.~Hays, P.~Perona, D.~Ramanan, P.~Dollár,
  C.~L. Zitnick, Microsoft coco: Common objects in context, in: D.~J. Fleet,
  T.~Pajdla, B.~Schiele, T.~Tuytelaars (Eds.), Computer Vision -- ECCV 2014,
  Vol. 8693 of Lecture Notes in Computer Science, Springer, 2014, pp. 740--755.
\newblock \href {https://doi.org/10.1007/978-3-319-10602-1_48}
  {\path{doi:10.1007/978-3-319-10602-1_48}}.

\bibitem{Rhodin2018}
H.~Rhodin, F.~Meyer, J.~Spörri, E.~Müller, V.~Constantin, P.~Fua,
  I.~Katircioglu, M.~Salzmann, Learning monocular 3d human pose estimation from
  multi-view images, in: 2018 IEEE/CVF Conference on Computer Vision and
  Pattern Recognition, 2018, pp. 8437--8446.

\bibitem{sapkota2025yolo26}
R.~Sapkota, R.~H. Cheppally, A.~Sharda, M.~Karkee, Yolo26: key architectural
  enhancements and performance benchmarking for real-time object detection,
  arXiv preprint arXiv:2509.25164 (2025).

\bibitem{carion2020end}
N.~Carion, F.~Massa, G.~Synnaeve, N.~Usunier, A.~Kirillov, S.~Zagoruyko,
  End-to-end object detection with transformers, in: Proceedings of the
  European Conference on Computer Vision (ECCV), Springer, 2020, pp. 213--229.

\bibitem{zhu2020deformable}
X.~Zhu, W.~Su, L.~Lu, B.~Li, X.~Wang, J.~Dai, Deformable detr: Deformable
  transformers for end-to-end object detection, arXiv preprint arXiv:2010.04159
  (2020).

\bibitem{zhang2022dino}
H.~Zhang, F.~Li, S.~Liu, L.~Zhang, H.~Su, J.~Zhu, L.~M. Ni, H.-Y. Shum, Dino:
  Detr with improved denoising anchor boxes for end-to-end object detection,
  arXiv preprint arXiv:2203.03605 (2022).

\bibitem{zhao2024detrs}
Y.~Zhao, W.~Lv, S.~Xu, J.~Wei, G.~Wang, Q.~Dang, Y.~Liu, J.~Chen, Detrs beat
  yolos on real-time object detection, in: Proceedings of the IEEE/CVF
  Conference on Computer Vision and Pattern Recognition (CVPR), 2024, pp.
  16965--16974.

\bibitem{jiang2023rtmpose}
T.~Jiang, P.~Lu, L.~Zhang, N.~Ma, R.~Han, C.~Lyu, Y.~Li, K.~Chen, Rtmpose:
  Real-time multi-person pose estimation based on mmpose, arXiv preprint
  arXiv:2303.07399 (2023).

\bibitem{yang2023effective}
Z.~Yang, A.~Zeng, C.~Yuan, Y.~Li, Effective whole-body pose estimation with
  two-stages distillation, in: 2023 IEEE/CVF International Conference on
  Computer Vision Workshops (ICCVW), IEEE, 2023, pp. 4212--4222.

\bibitem{sun2019deep}
K.~Sun, B.~Xiao, D.~Liu, J.~Wang, Deep high-resolution representation learning
  for human pose estimation, in: 2019 IEEE/CVF conference on computer vision
  and pattern recognition (CVPR), IEEE, 2019, pp. 5686--5696.

\bibitem{Zhao_2023_CVPR}
Q.~Zhao, C.~Zheng, M.~Liu, P.~Wang, C.~Chen, Poseformerv2: Exploring frequency
  domain for efficient and robust 3d human pose estimation, in: Proceedings of
  the IEEE/CVF Conference on Computer Vision and Pattern Recognition (CVPR),
  2023, pp. 8877--8886.

\bibitem{motionbert2022}
W.~Zhu, X.~Ma, Z.~Liu, L.~Liu, W.~Wu, Y.~Wang, Motionbert: A unified
  perspective on learning human motion representations, in: Proceedings of the
  IEEE/CVF International Conference on Computer Vision, 2023.

\bibitem{mehraban2024motionagformer}
S.~Mehraban, V.~Adeli, B.~Taati, Motionagformer: Enhancing 3d human pose
  estimation with a transformer-gcnformer network, in: 2024 IEEE/CVF Winter
  Conference on Applications of Computer Vision (WACV), IEEE, 2024, pp.
  6905--6915.

\bibitem{lea2017temporal}
C.~Lea, M.~D. Flynn, R.~Vidal, A.~Reiter, G.~D. Hager, Temporal convolutional
  networks for action segmentation and detection, in: proceedings of the IEEE
  Conference on Computer Vision and Pattern Recognition, 2017, pp. 156--165.

\bibitem{dey2017gate}
R.~Dey, F.~M. Salem, Gate-variants of gated recurrent unit (gru) neural
  networks, in: 2017 IEEE 60th international midwest symposium on circuits and
  systems (MWSCAS), IEEE, 2017, pp. 1597--1600.

\bibitem{he2016deep}
K.~He, X.~Zhang, S.~Ren, J.~Sun, Deep residual learning for image recognition,
  in: Proceedings of the IEEE conference on computer vision and pattern
  recognition, 2016, pp. 770--778.

\bibitem{tong2022videomae}
Z.~Tong, Y.~Song, J.~Wang, L.~Wang, Videomae: Masked autoencoders are
  data-efficient learners for self-supervised video pre-training, Advances in
  neural information processing systems 35 (2022) 10078--10093.

\bibitem{drawing_room}
{The Hallwyl Museum (Hallwylska museet)}, The great drawing room,
  \url{https://skfb.ly/6ypJL}, 3D model from Sketchfab, licensed under CC BY
  4.0.

\bibitem{bigazzi2022embodied}
R.~Bigazzi, F.~Landi, S.~Cascianelli, M.~Cornia, L.~Baraldi, R.~Cucchiara,
  Embodied navigation at the art gallery, in: International Conference on Image
  Analysis and Processing, Springer, 2022, pp. 739--750.

\bibitem{queisner2022volumetricor}
M.~Queisner, M.~Pogorzhelskiy, C.~Remde, J.~Pratschke, I.~M. Sauer,
  Volumetricor: A new approach to simulate surgical interventions in virtual
  reality for training and education, Surgical Innovation 29~(3) (2022)
  406--415.

\bibitem{mersus_office}
M.~Technologies, Mersus office, \url{https://skfb.ly/o7NFy}, 3D model,
  Sketchfab. Licensed under Creative Commons Attribution 4.0.

\bibitem{city_buildings}
abhayexe, Full gameready city buildings, \url{https://skfb.ly/p8Qzx}, 3D model
  from Sketchfab, licensed under CC BY 4.0.

\bibitem{via_gemini}
A.~S. (Spogna), Via s. gemini (viterbo - italy), \url{https://skfb.ly/6x7np},
  3D model from Sketchfab, licensed under CC BY 4.0.

\end{thebibliography}

\end{document}